# Computational thematics: Comparing algorithms for clustering the genres of literary fiction

Oleg Sobchuk[1,2], Artjoms Šeļa[3]

1 Max Planck Institute for Evolutionary Anthropology, Leipzig, Germany
2 Max Planck Institute for Geoanthropology, Jena, Germany
3 Institute of Polish Language, Polish Academy of Sciences, Krakow, Poland

## Abstract

What are the best methods of capturing thematic similarity between literary texts? Knowing the answer to this question would be useful for automatic clustering of book genres, or any other thematic grouping. This paper compares a variety of algorithms for unsupervised learning of thematic similarities between texts, which we call "computational thematics". These algorithms belong to three steps of analysis: text preprocessing, extraction of text features, and measuring distances between the lists of features. Each of these steps includes a variety of options. We test all the possible combinations of these options: every combination of algorithms is given a task to cluster a corpus of books belonging to four pre-tagged genres of fiction. This clustering is then validated against the "ground truth" genre labels. Such comparison of algorithms allows us to learn the best and the worst combinations for computational thematic analysis. To illustrate the sharp difference between the best and the worst methods, we then cluster 5000 random novels from the HathiTrust corpus of fiction.

**Keywords:** text mining, computational literary studies, genre, topic modeling

## Introduction

Computational literary studies have rapidly grown in prominence over the recent years. One of the most successful directions of inquiry within this domain, in terms of both methodological advances and empirical findings, has been computational stylometry, or computational stylistics: a discipline that develops algorithmic techniques for learning *stylistic similarities* between texts (Bories et al., 2023; Burrows, 1987; Eder et al., 2016). For this purpose, computational stylometrists extract linguistic features specifically associated with authorial style, or individual authorial habits. Often, these features are the most frequent words from the analyzed literary texts – they tend to be function words ("a", "the", "on", etc.) – to which various measures of similarity (e.g., Euclidean distance) are applied. The most common goal of computational stylistics is attributing the authorship of texts where it is disputed, like the authorship of Molière's plays (Cafiero & Camps, 2019), the Nobel Prize winning novel *And Quiet Flows the Don* (Iosifyan & Vlasov, 2020), or Shakespeare and Fletcher's play *Henry VIII* (Plecháč, 2021). Thanks to numerous systematic comparisons of various approaches to computational stylometry, we now have a fairly good idea of which procedures and textual features are the most effective ones – depending on the goal of stylometric analysis, the language of texts, or their genre (Evert et al., 2017; Neal et al., 2017; Plecháč et al., 2018).

At the same time, we lack such systematic comparisons in the research area that might be called "computational thematics": the study of *thematic similarities* between texts. (Thematic similarities: say, that novels A and B both tell a love story or have a "fantasy" setting.) Why is learning about thematic similarities important? Genre – a population of texts united by broad thematic similarities – fantasy, romance, science fiction, and the like – is a central notion in literary studies, necessary not only for categorizing and cataloging literary works, but also for the historical scholarship of literature. Genres are evolving populations of texts that emerge at certain moments of time, spread across the field of literary production, and then disappear in their original form – usually becoming stepping stones for subsequent genres (Fowler, 1971). For example, the genre of "classical" detective fiction crystallized in the 1890–1930s, and then gave birth to multiple other genres of crime fiction, such as "hardboiled crime fiction", "police procedural", "historical detective", and others (Symons, 1985). Studying the historical dynamics of genres – not only of literature, but also music or painting – is an important task of art history and sociology, and digital archives allow doing so on a much larger scale (Allison et al., 2011; Klimek et al., 2019; Sigaki et al., 2018). But to gain the most from this larger scale, we must determine the best, most reliable algorithms for detecting the thematic signal in books – similarly to how computational stylometrists have learnt the most effective algorithms for detecting the signal of authorship.

Quantitative analysis of genres usually takes one of these forms. The first one is *manual tagging* of books by genre or using datasets where such tagging has already been done via large crowdsourced efforts, like the data collected on the Goodreads website (Thelwall, 2019). This approach is prone to human bias, it is laborious and also based on the idea that the differences between genre populations are qualitative, not quantitative (e.g., certain book is either a "detective" or "romance", or both, but not 0.78 detective and 0.22 romance, which, we think, would be a more informative description). The second approach is an extension of manual tagging: *supervised machine learning* of book genres using a training dataset with manually tagged genres (Piper et al., 2021; Underwood, 2019). This approach has important strengths: it is easily scalable and it provides not qualitative but quantitative estimates of a book's belongingness to a genre. Still, it has a problem: it can only assign genre tags included in the training dataset, it cannot find new, unexpected book populations – which is an important component of the historical study of literature. The third approach is *unsupervised clustering* of genres: algorithmic detection of book populations based on their similarity to each other (Calvo Tello, 2021; Schöch, 2017). This approach is easily scalable, allows quantitative characterization of book genres, and does not require a training dataset with manually assigned tags, thus allowing to detect new, unexpected book populations. All these features of unsupervised clustering make it highly suitable for historical research, and this is why we will focus on it in this paper.

Unsupervised clustering can be conducted in a variety of ways. For example, texts can be lemmatized or not lemmatized; as text features, simple word frequencies can be used or some higher-level units, such as topics of a topic model; to measure the similarity between texts, a host of distance metrics can be applied. Hence, the question: what are the best computational methods for detecting thematic similarities in literary texts? This is the main question of this paper. To answer it, we will compare various combinations of (1) preprocessing (which, in this study, we will also call "thematic foregrounding"), (2) text features, and (3) the metrics used for measuring distance between features. To assess the effectiveness of these combinations, we use a tightly controlled corpus of four well-known genres – detective fiction, science fiction, fantasy, and romance – as our "ground truth" dataset. To illustrate the significant difference between the best and the worst combinations of algorithms for genre detection, we later cluster genres in a much larger corpus, containing 5000 works of fiction.

## Materials and Methods

### Data: The "ground truth" genres

Systematic research on computational stylistics is common, while research on computational thematics is still rare (Allison et al., 2011; Schöch, 2017; Šeļa et al., 2022; Underwood, 2016). Why? Computational stylistics has clear "ground truth" data against which various methods of text analysis can be compared: authorship. The methods of text analysis in computational stylistics (e.g., Delta distance or Manhattan distance) can be compared as to how well they perform in the task of classifying texts by their authorship. We write "ground truth" in quotes, as authorship is no more than a convenient proxy for stylistic similarity, and, as any proxy, it is imprecise. It assumes that texts written by the same author should be more similar to each other than texts written by different authors. However, we know many cases when the writing style of an author would evolve significantly over the span of their career, or would be deliberately manipulated (Brennan et al., 2012). Authorship as a proxy for "ground truth" is a simplification – but a very useful one.

The lack of a widely accepted "ground truth" proxy for thematic analysis leads to the comparisons of algorithms that are based on nothing more than subjective judgment (Egger & Yu, 2022). Such subjective judgment cannot lead us far: we need quantitative metrics of performance of different algorithms. For this, an imperfect "ground truth" is better than none at all. What could play the role of such an imperfect, but still useful, ground truth in computational thematics? At the moment, these are genre categories. They capture, to a different degree, thematic similarity between texts. To a different degree, as genres can be organized according to several principles, or "axes of categorization": e.g., they can be based on the similarity of storylines (adventure novel, crime novel, etc.), settings (historical novel, dystopian novel, etc.), emotions they evoke in readers (horror novel, humorous novel, etc.), or their target audience (e.g., young adult novels). It does seem that these various "axes of categorization" correlate: say, "young adult" novels are appreciated by young adults because they often have similar storylines or characters. Or, horror novels usually have a broad, but consistent, arsenal of themes and settings that are efficient at evoking pleasant fear in readers (like the classical Gothic setting). Still, some axes of genre categorization are probably better for comparing the methods of computational thematics than others. Genres defined by their plots or settings may provide a clearer thematic signal than genres defined by their target audience or evoked emotions.

We have assembled a tightly controlled corpus of four genres (50 texts in each) based on their plots and settings:

- Detective fiction (recurrent thematic elements: murder, detective, suspects, investigation)

- Fantasy fiction (recurrent elements: magic, imaginary creatures, quasi-medieval setting)
- Romance fiction (recurrent elements: affection, erotic scenes, love triangle plot)
- Science fiction (recurrent thematic elements: space, future, technology)

We took several precautions to remove potential confounds. First, these genres are situated on a similar level of abstraction: we are not comparing rough-grain categories (say, romance or science fiction) to fine-grain ones (historical romance or cyberpunk science fiction). Second, we limited the time span of the book publication year to a rather short period of 1950–1999: to make sure that our analysis is not affected too much by language change (which would inevitably happen if we compared, for example, 19th-century gothic novels to 20th-century science fiction). Third, each genre corpus has a similar number of authors (29–31 authors), each represented by 1–3 texts. Several examples of books in each genre are shown in **Table 1**. The complete list is in **Supplementary materials**. Before starting our analysis, we pre-registered this list on Open Science Framework's website (https://osf.io/rce2w/?view_only=16db492ab4464a4da53b1ef891416bd4).

| Genre | Examples |
|---|---|
| Detective fiction  | Josephine Tey, *The Daughter of Time*, 1951<br>Agatha Christie, *At Bertram's Hotel*, 1965<br>Colin Dexter, *Last Bus to Woodstock*, 1975<br>Peter Lovesey, *The False Inspector Dew*, 1982<br>Sue Grafton, *M is for Malice*, 1996 |
| Fantasy fiction  | J. R. R. Tolkien, *The Fellowship of the Ring*, 1954<br>Michael Moorcock, *Stormbringer*, 1965<br>Ursula K. Le Guin, *The Tombs of Atuan*, 1970<br>Terry Pratchett, *The Colour of Magic*, 1983<br>J. K. Rowling, *Harry Potter and the Philosopher's Stone*, 1997 |
| Romance fiction  | Barbara Cartland, *Love is the Enemy*, 1952<br>Jackie Collins, *The World is Full of Married Men*, 1968<br>Gordon Merrick, *The Lord Won't Mind*, 1970<br>Danielle Steel, *A Perfect Stranger*, 1981<br>Diana Gabaldon, *Outlander*, 1991 |
| Science fiction  | Robert A. Heinlein, *Double Star*, 1956<br>Arthur C. Clarke, *2001: A Space Odyssey*, 1968<br>Frank Herbert, *Children of Dune*, 1976<br>C. J. Cherryh, *Downbelow Station*, 1981<br>Kim Stanley Robinson, *Red Mars*, 1992 |

**Table 1**. Examples of books in each genre corpus (full list in **Supplementary materials**).

**Analysis: The race of algorithms**

To compare the methods of detecting thematic signal, we developed a workflow consisting of four steps – see **Figure 1**. Same as our corpus, all the detailed steps of the workflow were pre-registered.

**Step 1. Choosing a combination of thematic foregrounding, features, and distance**

As a first step, we choose a combination of (a) the level of thematic foregrounding, (b) the features of analysis, and (c) the measure of distance.

By *thematic foregrounding* (**Step 1a** on **Figure 1**) we mean the extent to which the thematic aspect of a text is highlighted (and the stylistic aspect – backdropped). With *weak* thematic foregrounding, only the most basic text preprocessing is done: lemmatizing words and removing 100 most frequent words (MFWs) – the most obvious carriers of strong stylistic signal. 100 MFWs roughly correspond to function words (or closed-class words) in English, routinely used in authorship attribution (Chung & Pennebaker, 2007; Stamatatos, 2009) beginning with the classical study of *Federalist Papers* (Mosteller & Wallace, 1963). With *medium* thematic foregrounding, in addition to lemmatizing, we also remove entities (named entities, proper names, etc.) using SpaCy tagger (Honnibal & Montani, 2017). Additionally, we perform part-of-speech tagging and remove all the words that are not nouns, verbs, adjectives, or adverbs, which are the most content-bearing parts of speech. With *strong* thematic foregrounding, in addition to all the steps of the medium foregrounding, we also apply lexical simplification. We simplify the vocabulary by replacing less frequent words with their more frequent synonyms – namely, we replace all words outside of 1000 MFWs with their more common semantic neighbors (out of 10 closest neighbors), with the help of pre-trained FastText model that includes 2 million words and is trained on English Wikipedia (Grave et al., 2018).

Then, we transform our pre-processed texts into lists of features (**Step 1b** on **Figure 1**). We vary both the type of features and the length of lists. We consider four types of features. The simplest features are most frequent words as used in the bag-of-words approach (1000, 5000, or 10,000 of them) – a common solution for thematic analysis in computational literary studies (Hughes et al., 2012; Underwood, 2019). The second type of features are topic probabilities generated with the Latent Dirichlet Allocation (LDA) algorithm (Blei et al., 2003) – another common choice (Jockers, 2013; Liu et al., 2021). LDA has several parameters that can influence results, such as the predefined $k$ of topics or the number of most frequent words used. Plus, a long text like a novel is too large for meaningful LDA topic modeling, and the typical solution is dividing the text into smaller chunks. We use an arbitrary chunk size of 1000 words. The third type of features are modules generated with weighted correlation network analysis, also known as weighted gene co-expression network analysis (WGCNA) – a method of dimensionality reduction that detects clusters (or "modules") in networks (Langfelder & Horvath, 2008). WGCNA is widely used in genetics (Bailey et al., 2016; Ramírez-González et al., 2018), but also showed promising results as a tool for topic modeling of fiction (Elliott, 2017). We used it with either 1000 or 5000 most frequent words. Typically, WGCNA is used without chunking data, but, since chunking leads to better results in LDA, we decided to try using WGCNA with and without chunking, with the chunk size of 1000 words. All the parameters of WGCNA were kept at defaults. Finally, as the fourth type of feature, we use document-level embeddings doc2vec (Lau & Baldwin, 2016; Le & Mikolov, 2014) that directly position documents in a latent semantic space defined by a pre-trained distributional language model – FastText (Grave et al., 2018). Document representations in doc2vec depend on the features of the underlying model: in our study, each document is embedded in 300 dimensions of the original model. Doc2vec and similar word embedding methods are increasingly used for assessing the similarity of documents (Dynomant et al., 2019; Kim et al., 2019; Pranjic et al., 2020). As a result of **Step 1b**, we obtain a document-term matrix formed of texts (rows) and features (columns).

Finally, we must learn the similarity between the texts represented with the chosen lists of features – by using some metric of distance (**Step 1c** on **Figure 1**). There exist a variety of metrics for this purpose: Euclidean, Manhattan, Delta, Cosine, Cosine Delta distances and Jensen–Shannon divergence (symmetrized Kullback–Leibler divergence) for features that are probability distributions (in our case, this can be done for LDA topics and bag-of-words features).

Variants of **Step 1a**, **1b**, and **1c**, can be assembled in numerous *combinations*. In our "race of algorithms", each combination is a competitor – and a potential winner. Say, we could choose a combination of weak thematic foregrounding, LDA topics with 50 topics on 5000 most frequent words, and Euclidean distance. Or, medium thematic foregrounding, simple bag-of-words with 10,000 most frequent words, and Jensen–Shannon divergence. Some of these combinations are researchers' favorites, while others are underdogs – used rarely, or not at all. Our goal is to map out the space of possible combinations – to empirically test how each combination performs in the task of detecting the thematic signal. In total there are 311 competing combinations.

## Step 2. Sampling for robust results

A potential problem with our experiment could be that some combinations might perform better or worse simply because they are more suitable to our corpus of novels – for whatever reason. To reduce the impact of individual novels in our corpus, we do cross-validation: instead of analyzing the corpus as a whole, we analyze smaller *samples* from the corpus multiple times. Each sample contains 120 novels: 30 books from each genre. Altogether, we perform the analysis for each combination on 100 samples. For each sample, all the models that require training – LDA, WGCNA, and doc2vec – are trained anew.

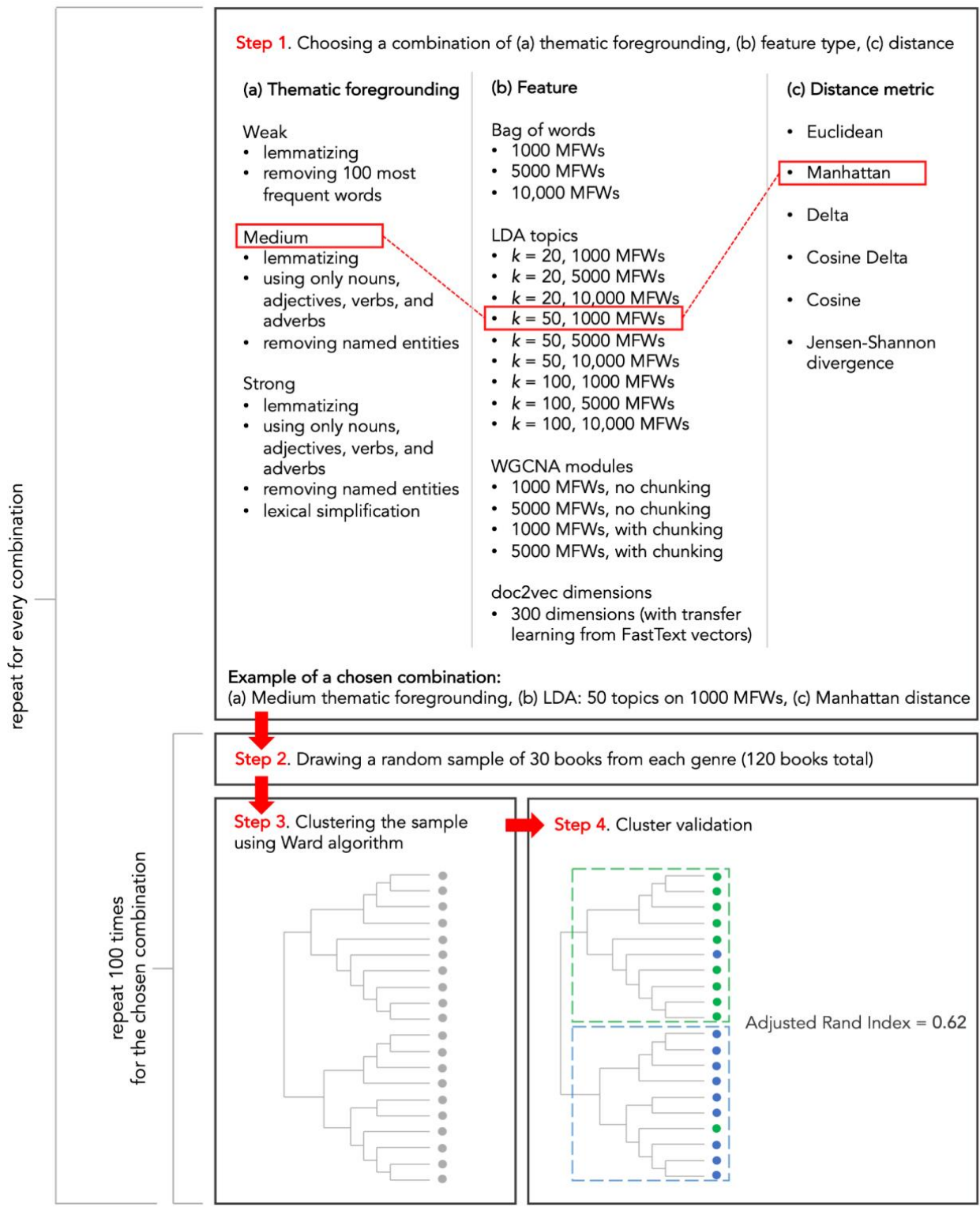

**Figure 1.** Four steps of the analysis. The workflow includes two loops. Big loop goes through various combinations of thematic control (Step 1a), feature type (1b), and distance metric (1c). For each such combination, a small loop is run: it randomly draws a genre-stratified sample of 120 novels (Step 2), clusters the novels using Ward algorithm (Step 3), and validates the clusters on the dendrogram using Adjusted Rand Index (Step 4). As a result of these four steps, each combination receives an ARI score: a score of its performance in detecting genres.

**Step 3. Clustering**

As a result of Step 2, we obtain a matrix of text distances. Then, we need to cluster the texts into groups – our automatically generated genre clusters, which we will later compare to the "true" clusters. For this, we could have used a variety of algorithms (e.g., k-means). We use hierarchical clustering with Ward's linkage (Ward, 1963): it clusters two items when resulting clusters maximize variance across the distance matrix. Despite being originally defined only for Euclidean distances, it was empirically shown that Ward's algorithm outperforms other linkage strategies in text-clustering tasks (Ochab et al., 2019). We assume that novels from four defined genres should roughly form four distinct clusters (as the similarity of texts within genre is greater than similarity of texts across genres). To obtain the groupings from a resulting tree we cut it vertically by the number of assumed clusters (which is 4).

**Step 4. Cluster validation**

How similar are our generated clusters to the "true" genre populations? To learn this, we compare the clusters generated by each chosen combination to the original genre labels. For this, we use a measure of cluster validation called the adjusted Rand index (ARI) **(Hubert & Arabie, 1985)**. ARI score of a particular combination shows how well this combination performs in the task of detecting genres – and thus, in picking the thematic signal. Steps 1–4 are performed for every combination, so that every combination receives its ARI score. In the end of the analysis, we obtain a dataset of 29,100 rows (291 combinations, each tested on 100 random samples).

## Results

**Figure 2** shows the average performance of all the combinations of thematic foregrounding, features, and distance metrics. Our first observation: the average ARI of the best performing algorithms ranges from 0.66 to 0.7, which is rather high for the complicated, noisy data that is literary fiction. This gives additional support to the idea that unsupervised clustering of fiction genres is possible. Even a cursory look at 10 best-performing combinations immediately reveals several trends. First, none of the top combinations have weak thematic foregrounding. Second, 6 out of 10 best-performing features are LDA topics. Third, 8 out of 10 distances on this list are Jensen–Shannon divergence.

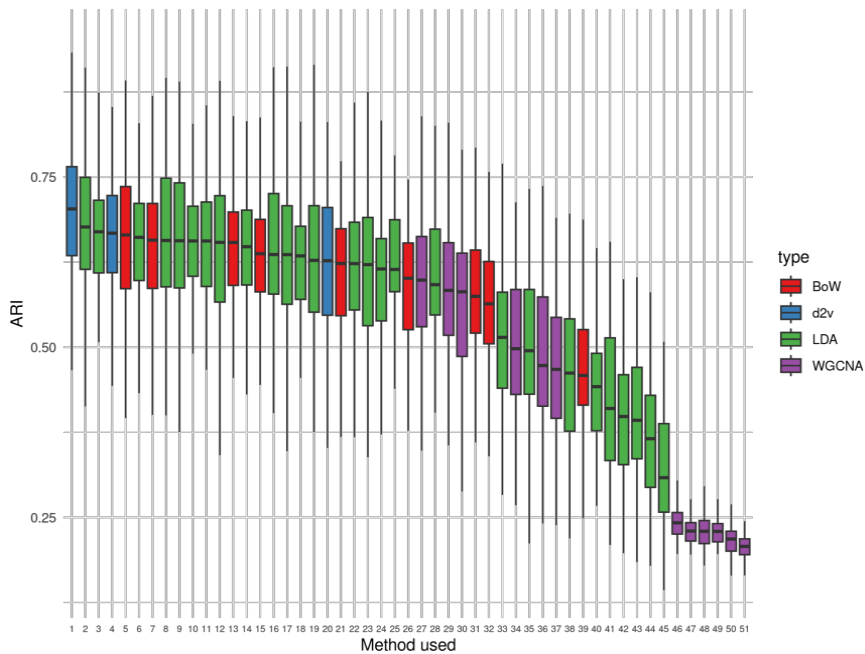

| Rank | Combination | | | Median ARI | Standard deviation |
|---|---|---|---|---|---|
| 1 | Strong foregr. | doc2vec (300 dimensions) | cosine | 0.703 | 0.092 |
| 2 | Strong foregr. | LDA (k=50, 5000 MFWs) | Jensen-Shannon | 0.677 | 0.107 |
| 3 | Strong foregr. | LDA (k=100, 1000 MFWs) | Jensen-Shannon | 0.670 | 0.086 |
| 4 | Medium foregr. | doc2vec (300 dimensions) | cosine | 0.668 | 0.093 |
| 5 | Strong foregr. | bag-of-words (10,000 MFWs) | Jensen-Shannon | 0.665 | 0.110 |
| 6 | Medium foregr. | LDA (k=50, 5000 MFWs) | Jensen-Shannon | 0.661 | 0.092 |
| 7 | Strong foregr. | bag-of-words (5000 MFWs) | Jensen-Shannon | 0.657 | 0.093 |
| 8 | Strong foregr. | LDA (k=20, 10,000 MFWs) | Jensen-Shannon | 0.657 | 0.114 |
| 9 | Strong foregr. | LDA (k=20, 5000 MFWs) | Jensen-Shannon | 0.656 | 0.116 |
| 10 | Strong foregr. | LDA (k=100, 5000 MFWs) | Jensen-Shannon | 0.656 | 0.074 |

**Figure 2.** Raw distributions of ARI scores for all the combinations of thematic foregrounding, feature type, and distance metric. Boxplots are colored by feature type. Numbers on the horizontal axis correspond to the names of combinations in the table to the right, showing 10 best-performing combinations (see all the combinations in **Supplement, Table S7**).

But how generalizable are these initial observations? How shall we learn the average "goodness" of a particular kind of thematic foregrounding, or a feature type, or a distance metric? To learn this, we need to control for their influence on each other, as well as for additional parameters, such as the number of most frequent words and chunking. Hence, we have constructed five Bayesian linear regression models (see **Supplement 5.1**). They answer questions about the performance of various combinations of thematic foregrounding, features, and distance metrics, helping us reach conclusions about the performance of individual steps of thematic analysis. All the results of this study are described in detail in **Supplement 5.1**. Below, we focus only on key findings.

**Conclusion 1. Thematic foregrounding improves genre clustering**

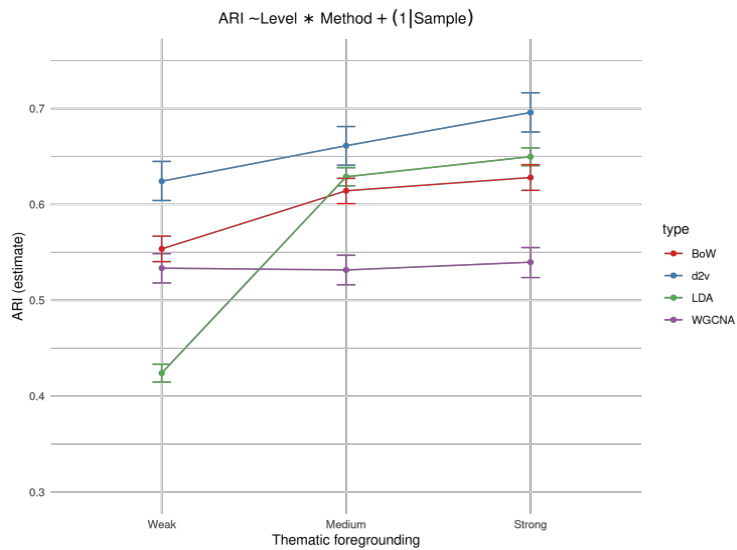

**Figure 3.** The effect of thematic foregrounding (weak, medium, or strong) on clustering genres, stratified by feature type.

The goal of thematic foregrounding was to highlight the contentful parts of the texts and to backdrop the stylistic parts. So, does larger thematic foregrounding improve genre recognition? As expected, we have found that low thematic foregrounding shows the worst performance across all four feature types (see **Figure 3**). For LDA and bag-of-words, it leads to drastically worse performance. At the same time, we do not see a large difference between the medium and the strong levels of thematic foregrounding. The major difference of the strong level of thematic foregrounding is the use of lexical simplification. This lexical simplification has not led to noticeable improvement of genre recognition. The gains of using strong thematic foregrounding for document embeddings, LDA and bag-of-words are marginal and inconsistent.

**Conclusion 2. Various feature types show similarly good performance**

Does the choice of feature type matter for the performance of genre clustering? We have found that almost all feature types can perform well. As shown on **Figure 2**, three out of four feature types – doc2vec, LDA, and bags of words – when used in certain combinations, can lead to almost equally good results. But how good are they on average? Figure 4 shows the posterior distributions of ARI for each type of features used in our analyses – in each case, for high level of thematic foregrounding.

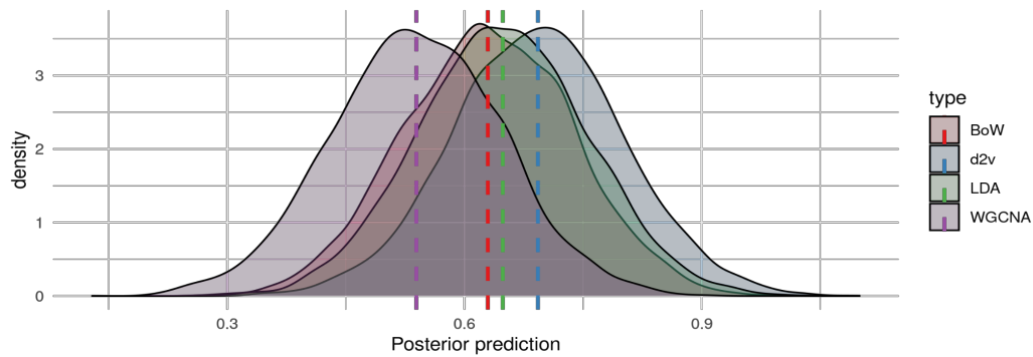

**Figure 4.** Posterior distributions of ARI scores for four feature types, at high level of thematic foregrounding.

As we see, doc2vec shows the best average performance, but this study has not experimented enough with using various other parameters of this feature type. It might be that another number of dimensions (e.g., 100 instead of 300) would worsen its performance. More research is needed to better understand the performance of doc2vec. LDA is the second best approach – and interestingly, the variation of parameters in LDA (such as $k$ of topics or $n$ of MFWs) does not increase the variance compared to doc2vec. Bag-of-words approach, despite being the simplest kind of feature, proves to be surprisingly good. It does not demonstrate the best performance, but it is not far behind doc2vec and LDA. At the same time, bags of words have a powerful advantage: simplicity. They are simpler to use and require fewer computational resources, meaning that in many cases they can still be a suitable choice for thematic analysis. Finally, WGCNA shows the worst ARI scores on average.

**Conclusion 3. The performance of LDA does not seem to depend on $k$ of topics and $n$ of most frequent words**

LDA modeling depends on parameters, namely $k$ of topics and $n$ of most frequent words, which should be decided, somewhat arbitrarily, before modeling. There exist algorithms for estimating the "good" number of topics, which help assessing how many topics are "too few" and how many are "too many" (Sbalchiero & Eder, 2020). In our study, however, we find no meaningful influence of either of these choices on learning the thematic signal (**Figure 5**). The single most important factor making a massive influence on the effectiveness of thematic classification is thematic foregrounding. Weak thematic foregrounding (in our case, only lemmatizing words and removing 100 most frequent words) proves to be a terrible choice that noticeably reduces ARI scores. Our study points towards the need for further systematic comparisons of various approaches to thematic foregrounding, as it seems to play a key role in the solid performance of LDA.

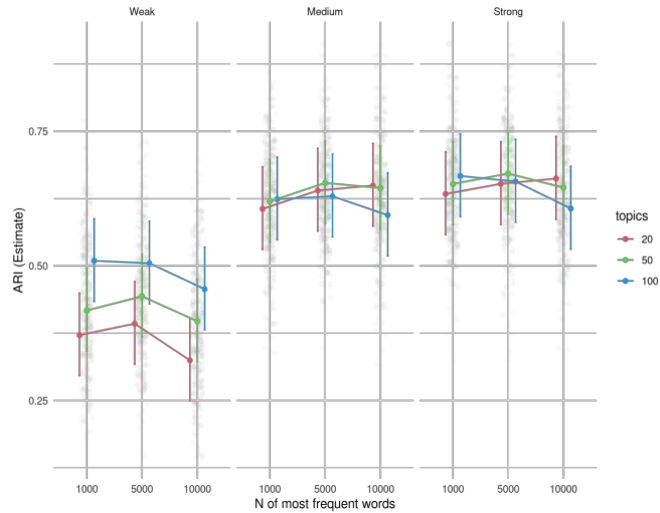

**Figure 5.** Posterior probabilities of the effects of k of topics on ARI, stratified by the level of thematic foregrounding and n of most frequent words used in LDA. Error bars show 95% credible intervals.

## Conclusion 4. Bag-of-words approach requires a balance of thematic foregrounding and *n* of most frequent words

Using bags of words as features is the simplest approach in thematic analysis, but still an effective one, as we have demonstrated. But how does one maximize the chances that bags of words perform well? We have varied two parameters in the bag-of-words approach: the level of thematic foregrounding and the number of MFWs used. **Figure 6** illustrates our findings: both these parameters influence the performance. Using 5000, instead of 1000, MFWs, drastically improves ARI scores. Similarly, using medium, instead of weak, thematic foregrounding, makes a big difference. At the same time, pushing these two parameters further – using 10,000 MFWs and strong thematic foregrounding – brings only marginal, if any, improvement in ARI scores.

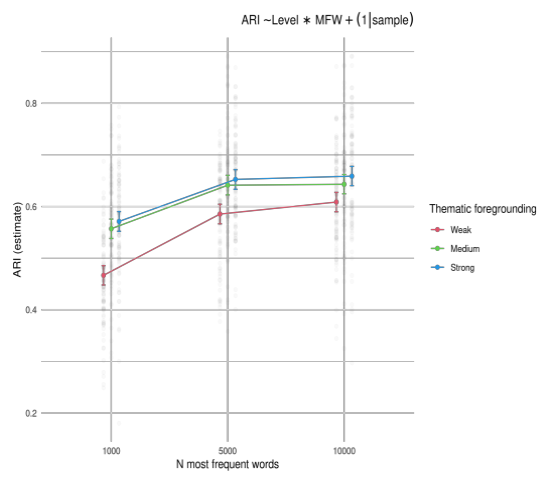

**Figure 6.** The influence of the number of most frequent words, used as text features, on learning the thematic signal, measured with ARI. There is a positive relationship between *n* of words and ARI, as well as between the level of thematic foregrounding and ARI. However, the middle parameter values of both (5000 MFWs and medium foregrounding) should be enough for most analyses.

**Conclusion 5. Jensen–Shannon divergence is the best distance metric for genre recognition, Euclidean – the worst**

Choosing the right distance metric is crucial for improving genre clustering. **Figure 7** shows the performance of various distances for each type of feature (note that Jensen–Shannon divergence, which was formulated for probability distributions, could not be applied to doc2vec dimensions and WGCNA module weights). For LDA and bag-of-words, Jensen–Shannon divergence is the best distance, with Delta and Manhattan distances being highly suitable too. For doc2vec, the choice of distance matters less. Interestingly, Euclidean distance is the worst-performing distance for LDA, bag-of-words, and WGCNA. This is an important, because this distance is often used in text analysis, also in combination with LDA (Jockers, 2013; Schöch, 2017; Underwood et al., 2022), while our study suggests that this distance should be avoided in computational thematic analysis. Cosine distance is also known to be useful for authorship attribution, when combined with bag-of-words as a feature type. At the same time, cosine distance is sometimes used to measure the distances between LDA topic probabilities, and our study shows that it is not the best combination.

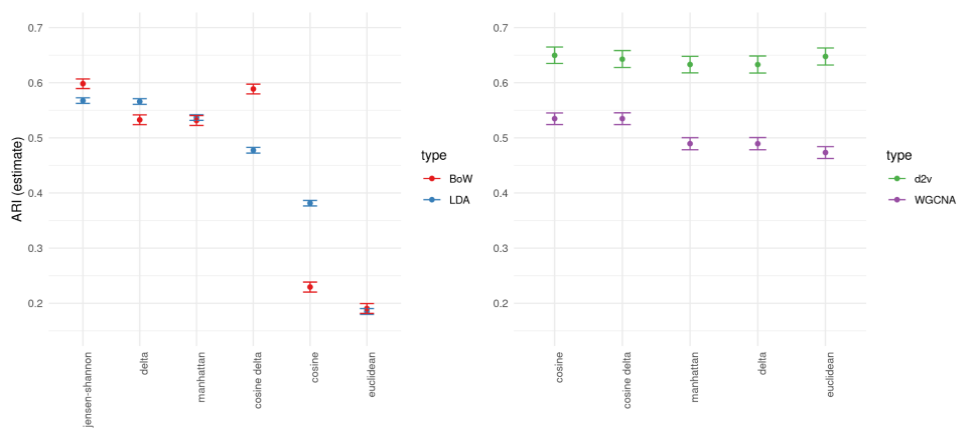

**Figure 7.** The influence of distance metrics on ARI scores, separately for each feature type. Note that Jensen–Shannon divergence could not be combined with WGCNA and doc2vec.

**Comparison of algorithms on a larger dataset**

How well does this advice apply to clustering other corpora, not just our corpus of 200 novels? A common problem in statistics and machine learning is overfitting: tailoring one's methods to a particular "sandbox" dataset, without making sure that these methods would work "in the wild". In our case, this means: would the same combinations of methods work well/poorly on other genres and other books than those included in our analysis? One precaution that we took to deal with overfitting was sampling from our genre corpus: instead of analyzing the full corpus just once, we analyzed smaller samples from it. But, additionally, it would be useful to compare the best-performing and the worst-performing methods against a much larger corpus of texts.

For this purpose, we use a sample of 5000 books of the NovelTM dataset of fiction, built from HathiTrust corpus (Underwood et al., 2020). Unlike our small corpus of four genres, these books do not have reliable genre tags, so we could not simply repeat our study on this corpus. Instead, we decided to inspect how a larger sample of our four genres (detective, fantasy, science fiction, and romance) would cluster in the HathiTrust corpus. For this, we included all the books in these four genres that we could easily identify (see **Supplement** for details) and seeded them into a random sample of 5000 works of fiction. Then we clustered all these books using two approaches: a particularly bad combination of methods for identifying genres (weak thematic foregrounding, bag-of-words with 5000 words, cosine distance) and a particularly good one (medium thematic foregrounding, LDA on 1000 words with 100 topics, clustered with Delta distance). The result, visualized with two UMAP projections (McInnes et al., 2018), is shown on **Figure 8**. One combination of methods resulted in a meaningful clustering, while the other – in chaos. However, this is only a first step towards further testing various algorithms of computational thematics "in the wild".

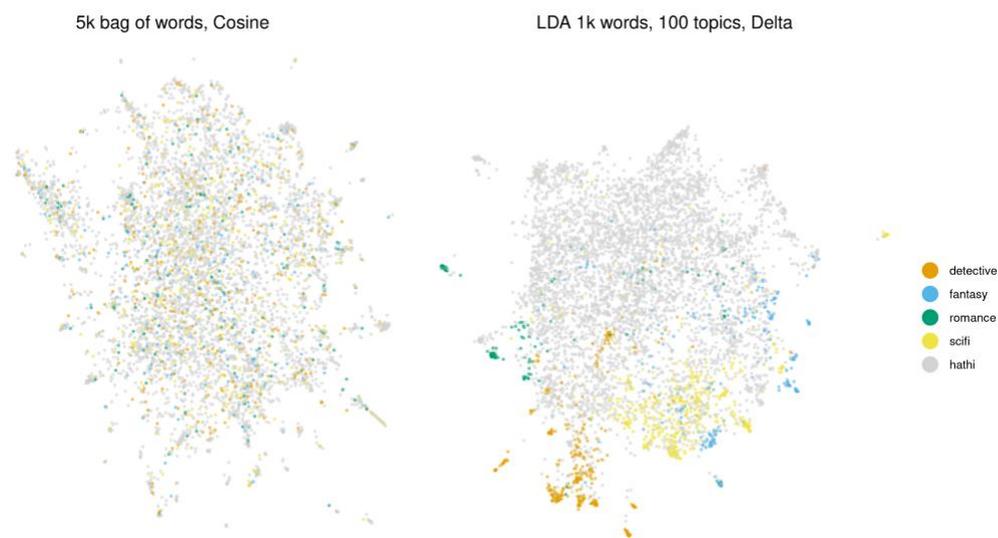

**Figure 8**. UMAP projections for a corpus consisting of 5,000 random novels from NovelTM HathiTrust corpus and all the novels all the authors included in the original corpus of four genres, found in NovelTM. Left-hand figure is clustered based on one of the worst-performing combinations, as found out by our study. Right-hand figure is based on one of the best-performing combinations.

## Discussion

This study aimed to answer the question: how good are various techniques of learning thematic similarities between works of fiction? In particular, how good are they at detecting genres – and are they good at all? For this, we tested various techniques of text mining, belonging to three consecutive steps of analysis: pre-processing, extraction of features, and measuring distances between the lists of features. We used four common genres of fiction as our "ground truth" data, including a tightly controlled sample of books. Our main finding is that unsupervised learning can be effectively used for detecting thematic similarities, but algorithms differ in their performance. Interestingly, the algorithms that are good for computational stylometry (and its most common task, authorship attribution) are not the same as those good for computational thematics. To give an example, one common approach to authorship attribution – using limited pre-processing, with a small number of most frequent words as features, and cosine distance – is one of the least accurate approaches for learning thematic similarities. How important are these differences in the real-world scenario, not limited to our small sample of books? To test this, we have contrasted one of the worst-performing combinations of algorithms, and one of the best-performing combinations, using a large sample of the HathiTrust corpus of books.

Systematic comparisons between various algorithms for computational thematic analysis will be key for a better understanding of which approaches work and which do not work – a requirement for assuring reliable results in the growing area of research which we suggest calling "computational thematics". Using a reliable set of algorithms for thematic analysis would allow tackling several large problems that remain not solved in the large-scale analysis of books. One such problem is creating better *genre tags* for systematizing large historical libraries of digitized texts. Manual genre tags in corpora such as HathiTrust are often missing or are highly inconsistent, which leads to attempts of using supervised machine learning, trained on manually tagged texts, to automatically learn the genres of books in the corpus overall. However, this approach, by design, allows capturing only the genres we already know about, and not the genres we do not know exist: "latent" genres. Unsupervised thematic analysis can be used for this task. Another important problem that unsupervised approaches to computational thematics may be good at is historical analysis of *literary evolution*. So far, we are lacking a comprehensive "map" of literary influences, based on the similarity of books. Such a map would allow creating a computational model of literary macroevolution, similar to phylogenetic trees (Bouckaert et al., 2012; Tehrani, 2013) or rooted phylogenetic networks (Neureiter et al., 2022; Youngblood et al., 2021) used in cultural evolution research of languages, music, or technologies. Having reliable unsupervised algorithms for measuring thematic similarities would be crucial for any historical models of this sort. Also, measuring thematic similarities may prove useful for creating *book recommendation* systems. Currently, book recommendation algorithms are mostly based on the analysis of user behavior: ratings or other forms of interaction (Duchen, 2022). Such methods are highly effective in the cases when user-generated data is abundant, like songs or brief videos. However, for longer content types, which take more time to consume and, the amount of user-generated data is much smaller. Improving the tools for content-based similarity detection in books would allow recommending books based on their content – as it is already happening to songs: projects such as Spotify's *Every Noise at Once* (https://everynoise.com/) combine user behavior data with the acoustic features of songs themselves to learn the similarity between songs and recommend them to listeners.

This study is a preliminary attempt at systematizing various approaches to computational thematics. More work is needed to further test the findings of this paper and to overcome its limitations. One apparent limitation is the concept of "ground truth" genres. It may be noted – rightly – that there are no "true" genres and that genre tags overall may not be the best approach for testing thematic similarities. As further steps we see using large scale user generated tags from Goodreads and similar websites as a proxy for "ground truth" similarity. Also, this study has certainly not exhausted all the possible techniques for text analysis that can be used for computational thematics. For example, a much wider testing of vector models, like doc2vec, but also BERTopic (Grootendorst, 2022) or Top2Vec (Angelov, 2020) is an obvious next step, or testing other network-based methods for community detection (Gerlach et al., 2018). Likewise, text simplification could have large potential for thematic analysis, it must be tested further. Possibly, the most straightforward way to test our findings would be attempting to replicate our results on other genre corpora, containing more books or other genres. Testing these methods on books in other languages is also critical. The approach taken in this paper offers a simple analytical pipeline – and we encourage other researchers to use it for testing all the various other computational approaches. Such a communal effort will be key for assuring robust results in the area of computational thematics.

## Competing interests

The author(s) declare no competing interests.

## Ethical approval

The study included no human or non-human participants, and thus requires no ethical approval.

## Supplementary Materials

R scripts used in our analysis, as well as reproducible Supplement with the detailed description of data and methods, can be found together with pre-registration documents on Open Science Framework's website [*anonymized link for peer-review*]: https://osf.io/rtvb6/?view_only=ea729efa0f504b9b8ea98e24ad60f6b9 Due to copyright law, we cannot share the corpus of books used in this study – instead we share document-term matrices based on the samples of this corpus.

# Computational thematics: Comparing algorithms for clustering the genres of literary fiction. Supplement

Anonymized

3/27/23

## Table of contents

# 1 Corpus summary

The corpus was constructed so that books roughly span the same time period across genres (Figure S1); also, each genre subcorpus does not include more than three books per author (Figure S2). The total number of authors contributing to each genre was also similar in each subcorpus.

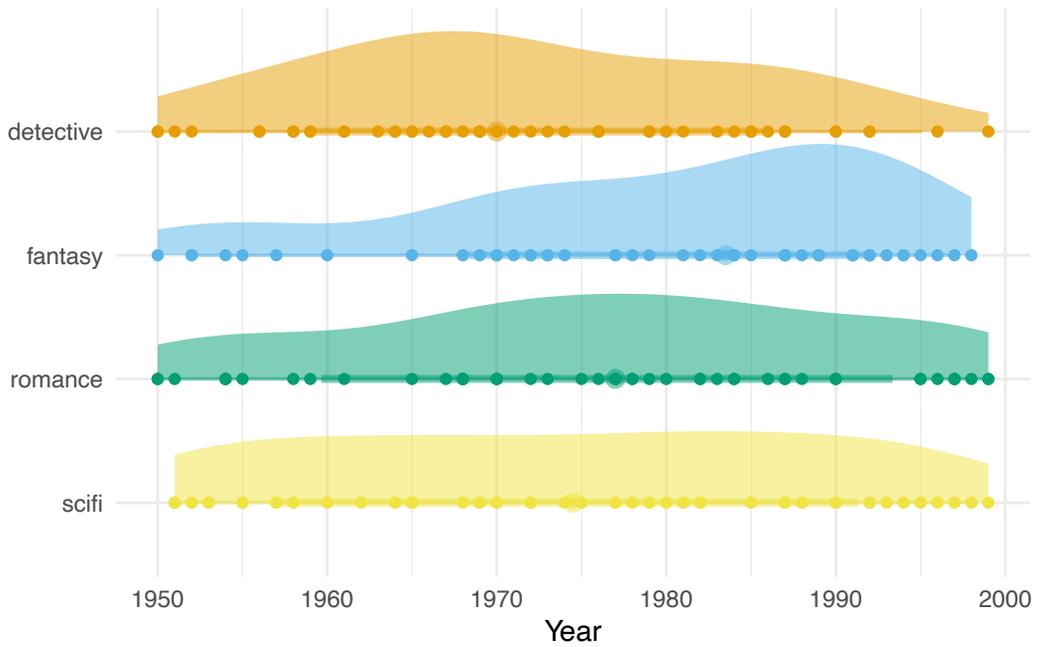

Figure S 1: Distribution of books through time within each genre.

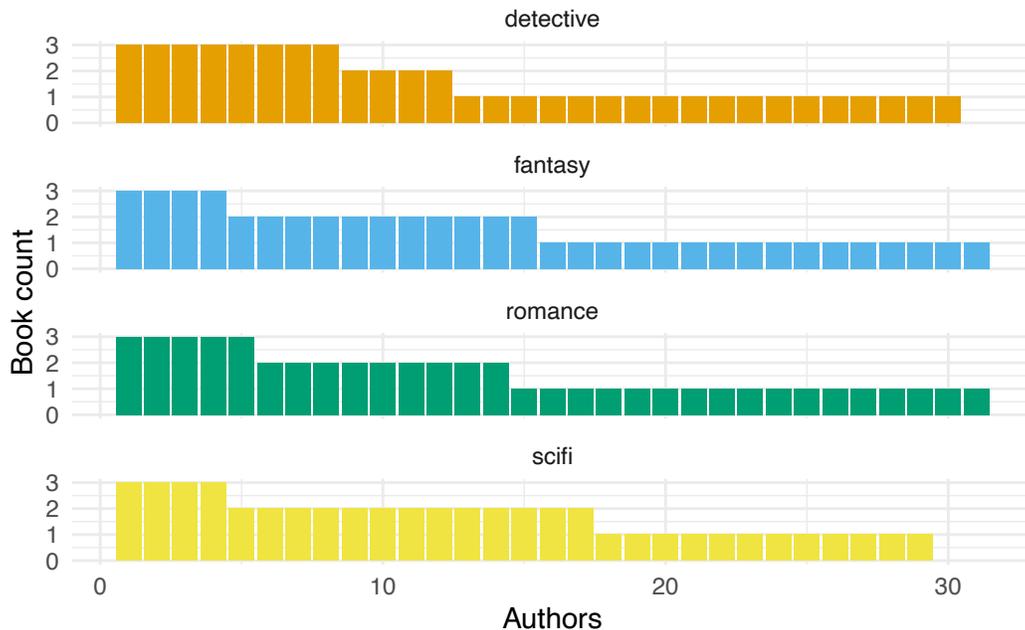

Figure S 2: Count of books belonging to a single author. Arranged by frequency rank.

## 2 Preprocessing

### 2.1 Thematic foregrounding: weak

At the first level of thematic foregrounding we remove 100 most frequent words (MFW) from analysis. 100 MFWs roughly correspond to function words (or closed-class words) in English (Chung & Pennebaker, 2007; Stamatatos, 2009) that are routinely used in authorship attribution starting from the classical study of *The Federalist Papers* (Mosteller & Wallace, 1963). MFWs can be removed to cheaply lower the impact of style (which heavily depends on grammar and syntactic differences) in favor of semantics and content.

### 2.2 Thematic foregrounding: medium

At the second level of thematic foregrounding words are pruned systematically based on morphology: we allow only nouns, adjectives, verbs and adverbs (auxiliary verbs are excluded too). We also remove entities and proper nouns, which might be specific to an author or a series of novels. Morphological tagging and named entity recognition was done with a basic spaCy language model for English due to its accessibility.

We did not use an external list of stopwords, since these lists are often arbitrary, can signficantly alter results, and are dominated by industry (specifically, information retrieval) standards. Lately, there is a tendency to minimize stopwords usage (Calvo Tello, 2021; Underwood et al., 2022), or to completely avoid them in the tasks like topic inference in a collection of documents (e.g. top2vec algorithm (Angelov, 2020)).

## 2.3 Thematic foregrounding: strong

The third level of thematic foregrounding includes steps from the medium foregrounding level and adds naive semantic simplification. We reduce the sparseness of feature space by turning less frequent words into more frequent words from similar semantic domains. We replace a word outside of 1000 MFWs with its closest semantic neighbor (out of the 10 closest neighbors) if this neighbor $n \in MFW$. To infer semantic similarity we use off-the-shelf FastText model (Mikolov et al., 2018) which includes 2M words and is trained on English Wikipedia, which provides a slice of 'modern' use of language. Again, this model is easily accessible and scalable to different tasks or languages.

Table S1 presents a random example of 20 semantic replacements.

As seen from examples, this lexical simplification can loosely sort target words by semantic domains represented by their more frequent semantic neighbors and, in some cases, clean original texts (`loove -> claim`). Noise is present, too, both from the domain-specific language of underlying word2vec model (`download -> free`) and the lack of context-based semantic disambiguation (`filmclip -> song`).

Finally, Figure S3 shows the filtering effects which different pre-processing strategies have on the corpus. The largest drop in word type diversity, predictably, happens after morphological filtering at medium thematic foregrounding; our naive lexical simplification allows removing another 5% of word types, but preserving the amount of tokens.

## 3 Features

### 3.1 Bag of words

A classic multivariate representation of texts as bags of words. We follow a stylometric tradition that assumes any weighting would also be part of the distance measure (e.g. Burrows' Delta is scaled Manhattan distance: see more about scaling features and vector length normalization in Evert et al. (2017)), so we only transform word frequencies to relative frequencies, ultimately dealing with a text as a probability distribution over an ordered set of words (arranged by their frequency) and defined by MFW cut-off. Different weightening techniques are widely used in information retrieval (TF-IDF, logarithmic transformation, etc.), but are more suitable as an Table S 1: A random sample of 20 'simplified' words. 'Source' column holds original words.

| source | replacement |
|--------|-------------|
| sundress | skirt |
| download | free |
| unlease | destroy |
| recycling | waste |
| organically | grow |
| trident | sword |
| snowed | snow |
| redeye | flight |
| cardholder | card |
| generate | create |
| struggling | struggle |
| dab | rub |
| houseparty | party |
| looove | love |
| grandmaster | master |
| filmclip | song |
| vantage | view |
| endeavor | effort |
| framing | frame |
| files | file |

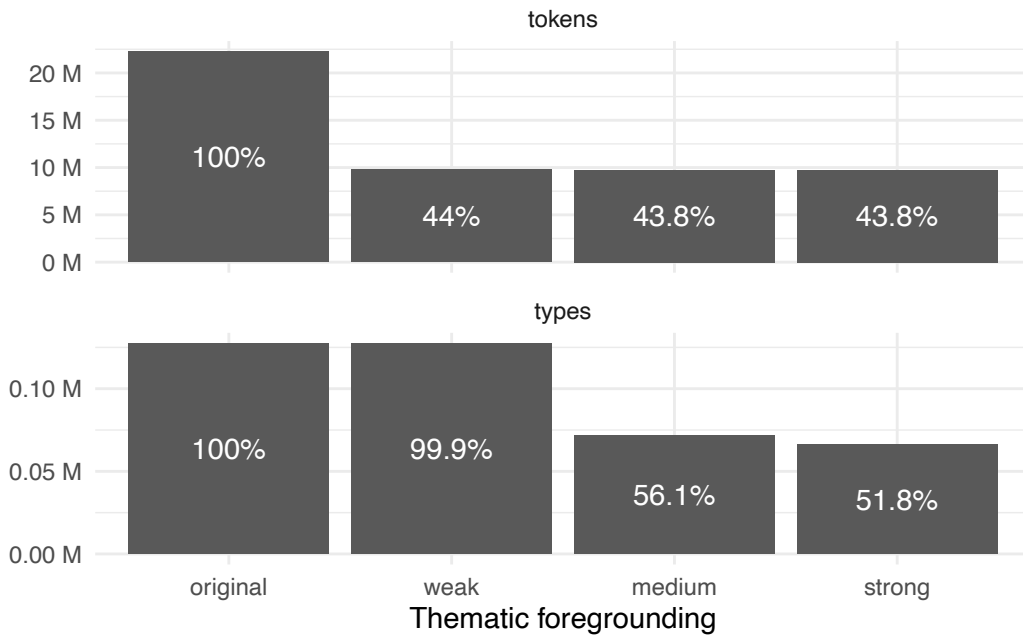

Figure S 3: Filtering effects on corpus. Percentages are given relative to the 'original' corpus size.

input to supervised learning (Calvo Tello, 2021) and might interfere with distance calculations (e.g. transforming original probabilities to something else).

### 3.2 Topic probabilities (LDA)

Latent Dirichlet Allocation (**blei__latent__2003?**) is the most widely used probabilistic algorithm of topic modeling, that still performs competitively with newer methods (Harrando et al., 2021). LDA infers groups of words (topics) based on their co-occurence in documents. Because LDA is generative, we can in turn represent each document as a probability distribution of topics. Compact lexical representation also makes the feature space more interpretable. We use `topicmodels` LDA implementation in R (Grün & Hornik, 2011). We vary parameters $k$ (number of topics) and MFW used. We do leave out hyperparameters *alpha* and *delta* at 0.1 default and do not rely on coherence/perplexity measures of a model, since we do not aim to fine-tune the LDA to a particular corpus; there is also empirical evidence that perceived LDA performance does not completely align with validation measures ((Hoyle et al., 2021); see also (Antoniak, 2022) for a summary of research on LDA performance).

An important pre-processing step for LDA is chunking of texts. A complete novel is too large of a context for inferring topics: too many words co-occur in large documents with many other words. Thus, instead of representing each novel as a bag of words, we represent it as many smaller bags of words from consecutive parts (chunks) before training an LDA. We use an arbitrary chunk size of 1000, but other structural cues (paragraphs, pages, chapters) for chunking might also be a good idea. We aggregate the probabilities from these smaller documents back to a single novel by taking an average of probability distributions (a centroid).

Table S 2: A sample of most probable words in LDA topics. Underlying model: full corpus, LDA, medium thematic foregrounding, 100 topics, 1000 MFWs.

| topic | terms |
|---|---|
| 2 | ship captain space sea leave time control war send command |
| 3 | hand eye head smile sit nod shake stand hold voice |
| 4 | father mother son daughter family child sister die live home |
| 7 | answer question speak doctor word call talk moment voice reply |
| 8 | feel eye smile walk moment sit suddenly hand slowly |
| 11 | human world planet life time system people space race war |
| 12 | dog bone animal bird mouth leg stick foot head eat |
| 15 | magic creature power castle change animal form head tree human |
| 16 | dragon fire wing fly eye head hold time land shoulder |
| 19 | time reach feel moment start completely hope system begin mind |

Table S2 demonstrates a sample of topics (10 most probable words per topic) in a model that is built using texts at medium level of thematic foregrounding, 100 topics, document-term matrix (DTM) was cut at 1000 MFWs. Topics clearly capture thematic groups like locations and settings, and are often linked to actions and relationships.

## 3.3 Module's weights (WGCNA)

Weighted gene correlation network analysis (WGCNA) is similar to LDA, but comes from a different research field: genetics. Some point out its promising features for text analysis, like relative independence from high-frequency function words (Elliott, 2017). WGCNA has one advantage over LDA: there is no need to guess the optimal number of topics, as WGCNA "modules" are determined automatically based on network of similarity in behavior between traits. Internally, WGCNA already relies on hierarchical clustering to derive modules that describe the variation in individuals/documents and could be greedy in reducing the word behavior in distinct genres to only one or two modules, especially if the analyzed texts have been chunked.

An example of this behavior from one of the sampling runs (120 novels) is presented on Figure S4. WGCNA was run on chunked novels, with medium thematic foregrounding, 5000 MFWs. We use implementation of the algorithm by Langfelder & Horvath (2008). The algorithm derived only one module of words that show almost perfectly opposite expression in detective and fantasy fiction, to no surprise: these genres are the easiest to distinguish. However, one module is too greedy when it comes to clustering: romance and sci-fi will be just mixed into two other distinct genres (and share more similarity to detectives than to fantasy).

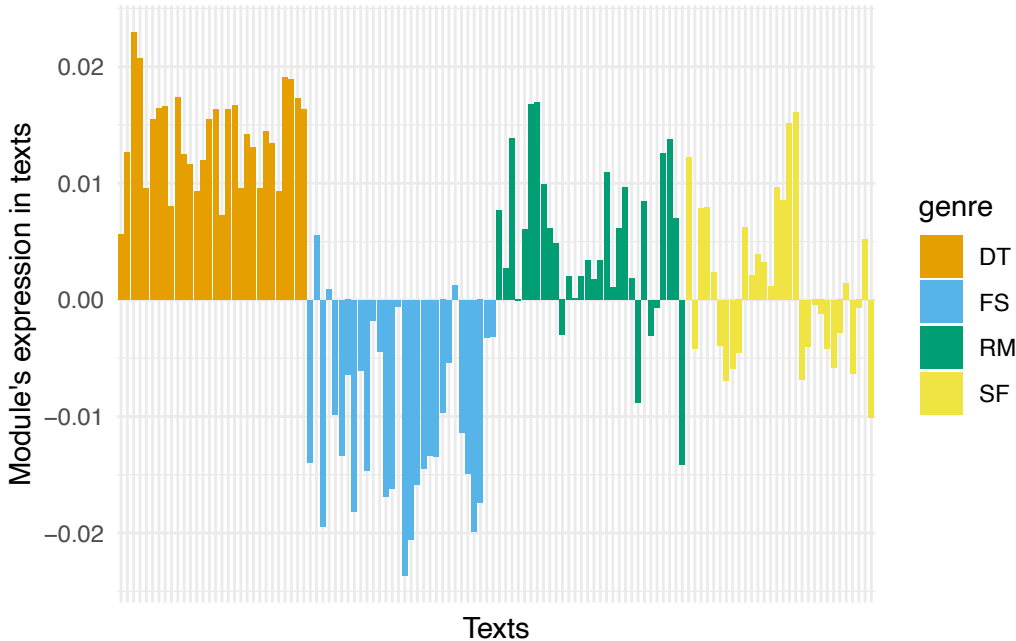

Figure S 4: Single-module problem in WGCNA

To find the most defining words for this global module we use a connectivity measure: in this case, these are words with the highest positive correlation to an "eigengene", which is a joint expression of a module in samples / documents. Figure S5 shows 20 most correlated words to a module from the same sample as on Figure S4. It is quite clear that these are words from a police procedural universe, and, more generally, these are words of a 'modern-like' urban setting, which also explains this module's expression in romance and science fiction. Conversely, the most inversely correlated words point to open spaces of adventure, magic and medieval attributes.

To give an example of WGCNA producing several meaningful distinct modules akin to LDA topics, we can use another model without chunking (medium thematic foregrounding, without chunking, 5000 MFWs). Examples of 10 most closely correlated words to a sample of modules are presented on Table S3. Unsurprisingly, removing chunking makes modules closely associated with specific books.

### 3.4 Document embeddings (doc2vec)

Doc2vec directly embeds documents into a latent semantic space that is defined by a distributional language model. In the end, each document is represented as a vector in this

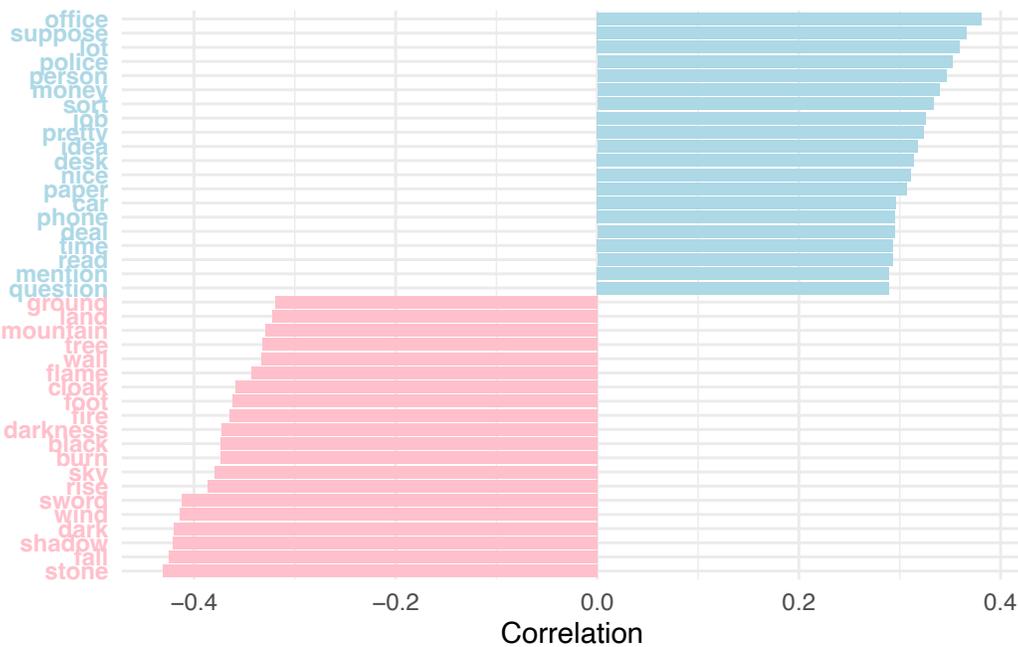

Figure S 5: Single-module WGCNA model: 5000 MFWs, medium thematic foregrounding, 1000-word chunks. Word correlations to the module's eigengene, 20 most and 20 less correlated words.

Table S 3: Words associated with WGCNA modules, random selection. 500 MFWs, medium thematic foregrounding, no chunking.

| module | words |
|---|---|
| 1 | bugger fighter commander formation strategy practice maneuver video bunk enemy |
| 4 | nuclear crisis mayor empire tech scientific science trader policy navy |
| 9 | dimension magician disguise assassin kid demon grumble flagon terrific mumble |
| 10 | hairy star meadow unicorn chain stall caravan gap innkeeper wax |
| 12 | sellsword tyrion Arya godswood maester raven eunuch ranger direwolf knight |
| 16 | menion mystic flick beyond sentry awesome massive attacker terrain quickly |
| 19 | oblige beg daresay countenance acquaint contrive disposition lordship shocking fashionable |
| 24 | laird iain elder outsider topic announce nudge blurt agreement argue |
| 25 | camera cowboy truck bridge vest vegetable brandy magazine corn bracelet |
| 26 | runciter talent organization anti spray employee commercial tv anyhow elevator |

*N*-dimensional space (depends on the underlying model). Again, we embed each novel split into chunks, in order to capture semantic variation on a small scale and then average the vectors in the space to get a single-vector-per-novel representation (a centroid of chunk vectors). We chose to use the chunks of 800 words and a pretrained FastText model (Mikolov et al., 2018) for vector representation of word semantics (300 dimensions, 2M words, trained on Wikipedia) and doc2vec implementation that allows fine-tuning and follows Angelov's alogrithm (2020).

Figure S6 shows UMAP projections of averaged novel vectors.

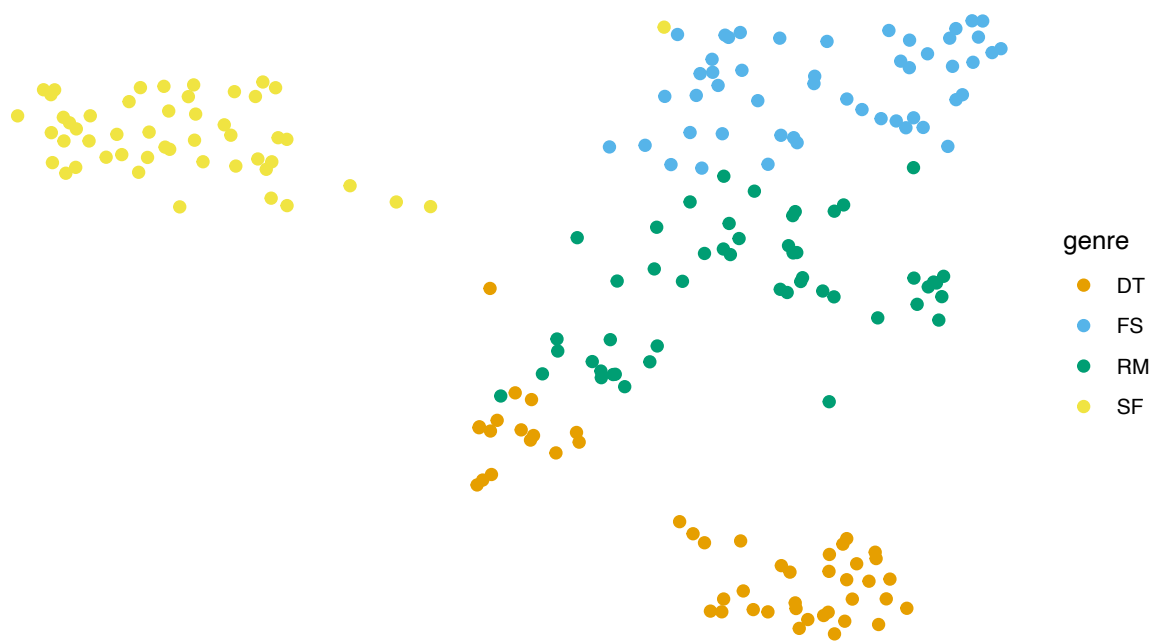

Figure S 6: UMAP projection of novel embedding 'centroids': average vectors of 800-words chunks; doc2vec, medium thematic foregrounding.

# 4 Clustering and validation

## 4.1 Distance measures

We infer similarity between novels by calculating pairwise distances between representations / vectors. We test several classic distances that are used for measuring text similarity (and are widely used in stylometry): Euclidean, Manhattan, Burrows' delta (scaled Manhattan), cosine, cosine delta (scaled cosine) and Jensen-Shannon divergence.

**Euclidean.** Square root of the squared pairwise differences in features

$$Euc(A, B) = \sqrt{\sum (A_i - B_i)^2}$$

**Manhattan**. Sum of the pairwise differences in dimensions (cityblock distance)

$$Man(A, B) = \sum (A_i - B_i)$$

**Burrows' delta**. Sum of the pairwise differences in scaled dimensions, normalized by vector length (Burrows, 2002). $z(A_i)$ is scaled and centered variable $_i$ in text $A$.

$$\Delta(A, B) = \frac{\sum z(A_i) - z(B_i)}{N}$$

**Cosine**. 1 - cosine similarity.

$$cos(A, B) = 1 - \frac{\sum A_i B_i}{\sqrt{\sum A_i^2} \sqrt{\sum B_i^2}}$$

**Cosine delta**. Same as cosine, but features are scaled (Evert et al., 2017).

**Jensen-Shannon Divergence**. Symmetrized Kullback-Leibler divergence. Not meaningful for feature vectors that are not probability distribution, but weights (e.g. WGCNA, doc2vec).

$$JSD(A, B) = 0.5 * (A - B) * (logit(A) - logit(B))$$

## 4.2 Clustering

In principle any other clustering algorithm could have been used (e.g. k-means). We use hierarchical clustering (Ward's linkage that pairs items when it minimizes within-cluster variance). Despite being originally defined only for Euclidean distances, it was empirically shown that Ward's algorithm outperforms other linkage strategies in text-clustering tasks (Ochab et al., 2019).

We assume that novels from four defined genres should roughly form four distinct clusters (similarity of texts within genre is greater than similarity of texts across genres). To obtain the groupings from a resulting tree we cut it vertically by the number of assumed clusters (k=4). Then we compare resulting classes to ideal clustering using Adjusted Rand Index (similar usages for unsupervised clutering with literary texts: (Cafiero & Camps, 2019; Šeļa et al., 2022)). ARI takes values between 1 and 0, where 1 would be a perfect classification and 0 would mean clustering not better than random.

## 4.3 Dendrogram of all novels

To provide an example of clustering performance, we build a dendrogram for all the novels in four genres (Figure S7). Underlying features are document embeddings at the medium level of thematic foregrounding and we use cosine distance for dissimilarity calculation. Colors of the branches are based on majority of genre neighbors. Adjusted Rand index of the tree presented below is 0.786.

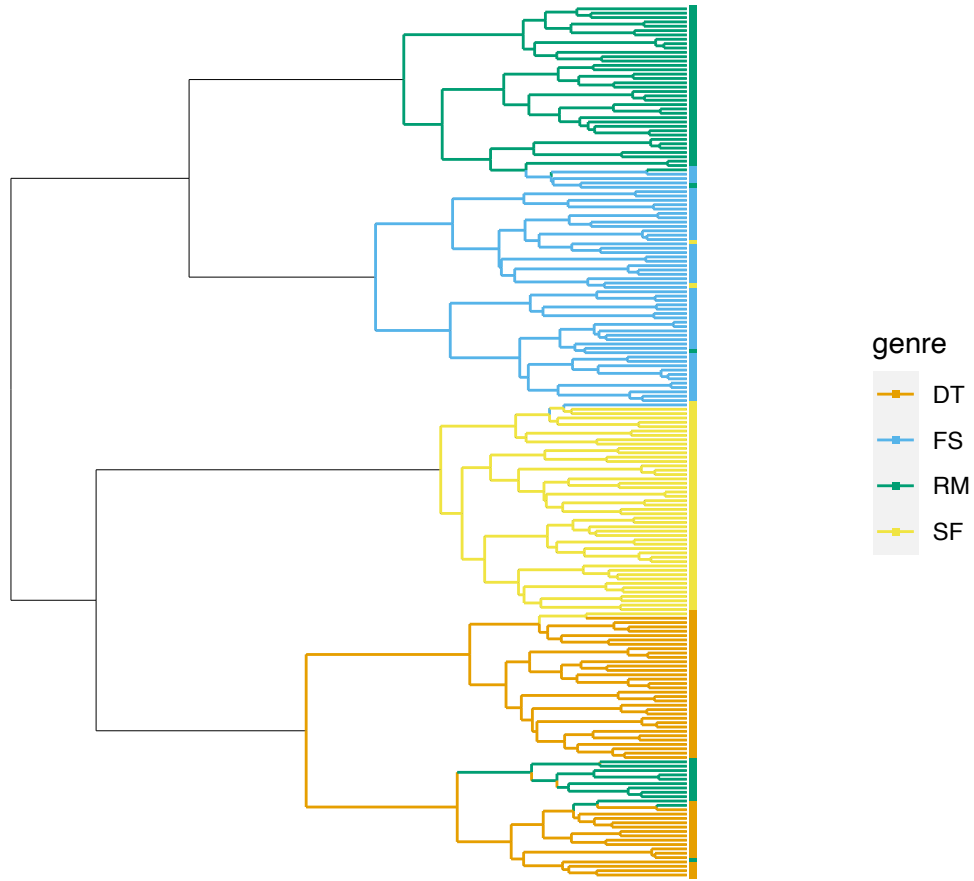

Figure S 7: Hierarchical clustering of the full corpus, doc2vec, medium thematic foregrounding, Ward's linkage. Clusters are colored by the dominant genre, up until a tree is cut to four major clusters.

Table S 4: Confusion matrix of clustering performance per genre.

|  | 1 | 3 | 2 | 4 |
|---|---|---|---|---|
| DT | 0.85 | 0.02 | 0.07 | 0.06 |
| FS | 0.07 | 0.77 | 0.05 | 0.11 |
| RM | 0.30 | 0.12 | 0.53 | 0.04 |
| SF | 0.13 | 0.24 | 0.07 | 0.55 |

Table S 5: Confusion matrix of clustering performance per genre (distances are filtered).

|  | 2 | 3 | 4 | 1 |
|---|---|---|---|---|
| DT | 0.90 | 0.01 | 0.06 | 0.03 |
| FS | 0.02 | 0.84 | 0.03 | 0.11 |
| RM | 0.27 | 0.06 | 0.64 | 0.02 |
| SF | 0.04 | 0.16 | 0.03 | 0.77 |

## 4.4 Confusion matrices

As seen from several figures above (S6, S7), genres differ in clustering consistency: detectives and fantasy books group together better than science fiction and romance. To address this difference in behavior we create a confusion matrix, based on all 100 cross-validation runs, which shows a dispersion of books across four clusters. Since this is not a supervised classification, a confusion matrix requires some heuristics to determine which clusters correspond to which genres in each clustering tree and can only show approximate results (we assume a cluster to be the 'detective' cluster if majority of books in this cluster are detectives).

This confusion matrix presents a total share of labeled novels that end up in different clusters across 29100 confusion matrices (100 samples, 291 clustering rounds in each). As usual, in the case of a pefect clustering, the diagonal of the matrix would contain "1". As expected, we see that the most diffused genres are romance (often grouped with detectives, 30% of hits) and science fiction (often grouped with fantasy, 24% of hits).

However, not all the methods summarized in the matrix above are equal and some distance measures (like Euclidean for bag of words) are 'bad choices' by default. To trim the matrix a little, we can follow the strategy that we also employ for modeling: use only well-suited distance measure for each method and remove chunked WGCNA, which proved to be a poor choice for thematic clustering.

Now the "good" clustering numbers are higher, but the difference between romance and science fiction becomes more pronounced. Comparatively, romance tends to form much more diffused clusters than science fiction (this tendency is visible on Figure S6).

Are different methods resulting in different sensitivity to genres and cluster formation? Figure S8 present a breakdown of the confusion matrix by document representation method.

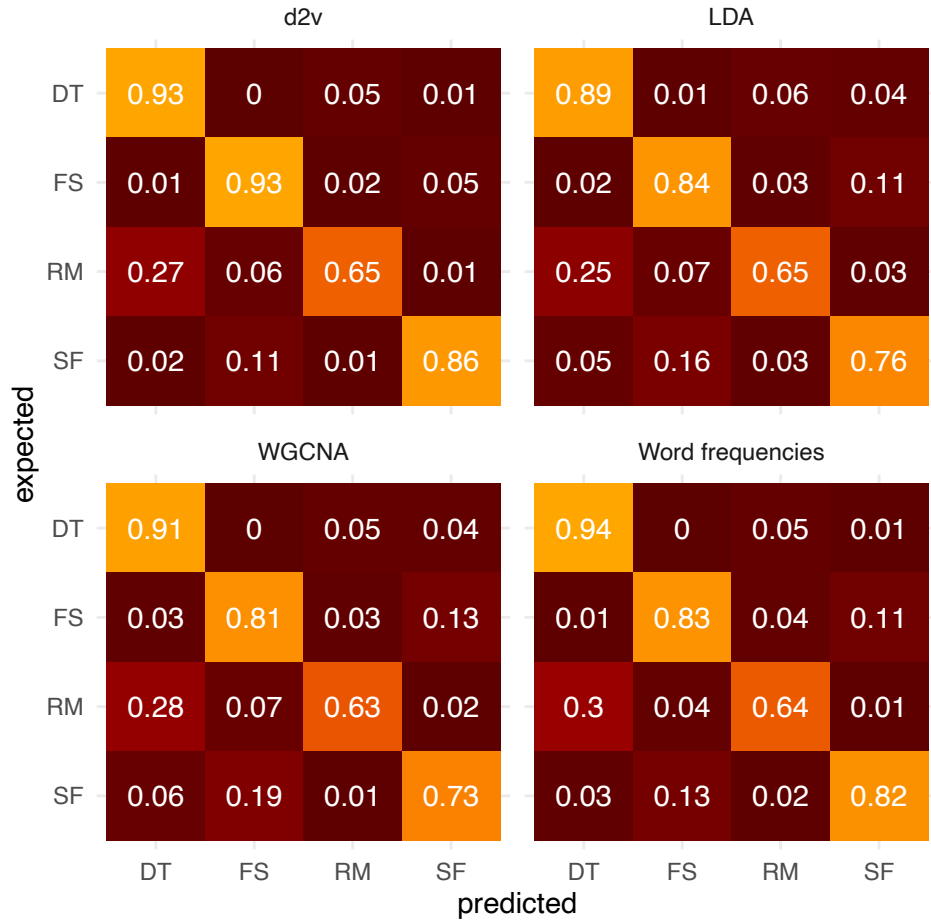

Figure S 8: Confusion matrices by different feature types.

Overall the same pattern holds across all methods, which is to be expected: they all rely on the same lexical frequency-based information. There is an advantage for d2v since it uses an external representations of word co-occurence based on a very large corpus, but higher numbers for doc2vec compared to LDA also should not be treated at face value, since it has fewer degrees of freedom (and, as a result, fewer ways to fail): doc2vec was used only in 3 different combinations per sample, while, for instance, LDA was used in 27.

# 5 Analysis

## 5.1 Linear models

Figure S9 shows the overall distribution of ARI values, with and without chunked WGCNA option. Value concentration on zero comes from unsuited distance choices that are inadequate for a given feature space (e.g. Euclidean for bag of words). When the data is filtered by better performing distances, distribution is not zero-inflated (see Figure S12).

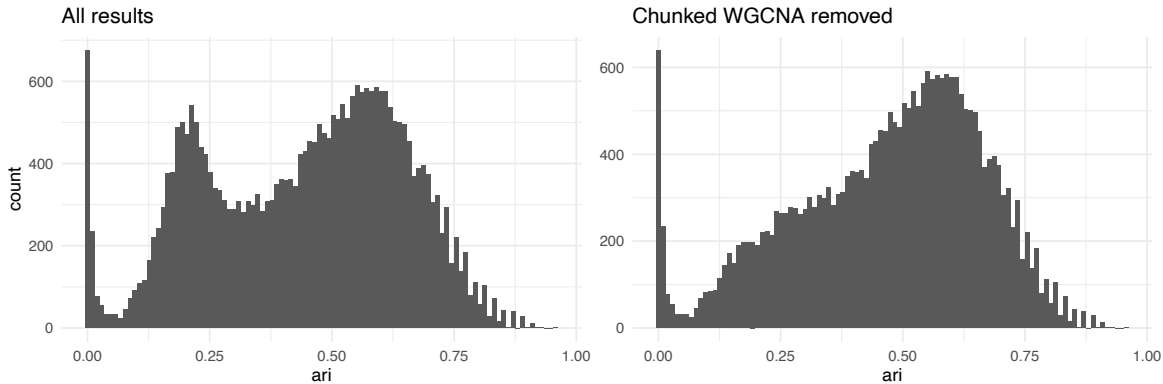

Figure S 9: Distribution of ARI values.

Alongside distance calculation and hierarchical clustering, we ran $k$-means clustering (with $k = 4$), but its average performance in separating books in four clusters (Figure S10), as measured by ARI, was way worse.

### 5.1.1 Distance selection

To simplify inference we will deal with results that were obtained with *suited* distance measure per each feature type. We select only the best performing distance measure per used feature. This is done to remove the factor of distances altogether and to equalize model's chances for comparison. There is no real reason to lump results from different distance measures together, since different data (e.g. probabilities vs. feature weights) has different sensitivity to distance selection, while some distances were not measured for some feature types (JSD for WGCNA and doc2vec).

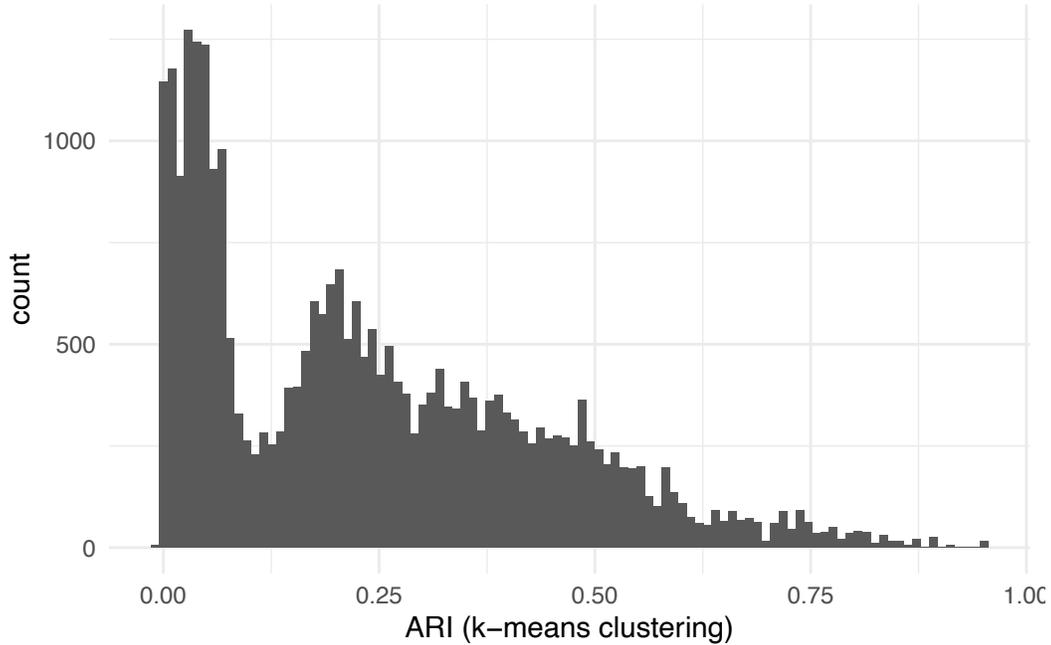

Figure S 10: Distribution of ARI values (k-means)

To choose the suited distances we fit a simple model `ari ~ 1 + feature*distance` to get estimates for each distance measure performance with each feature type (Figure S11). All further models were built using distances with highest posterior averages.

The list of chosen distances with estimates:

1. BoW: Jensen-Shannon (0.58)

2. LDA: Jensen-Shannon (0.57)

3. WGCNA: cosine delta (0.53)

4. doc2vec: cosine (0.65)

Figure S12 shows the ARI distribution for the filtered set of distances.

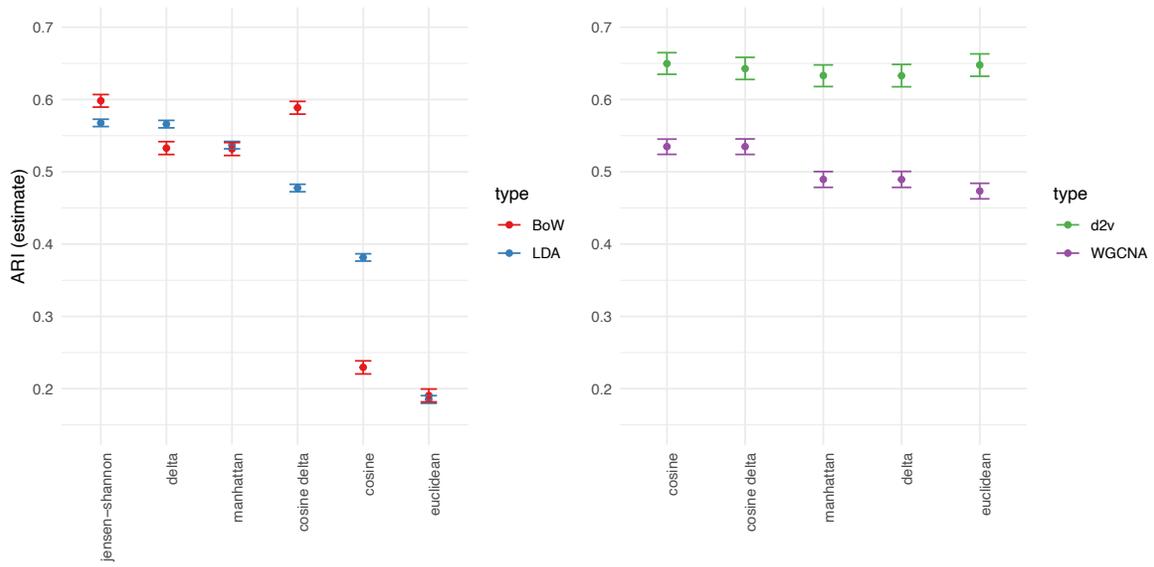

Figure S 11: Posterior predictions for distance performance across methods.

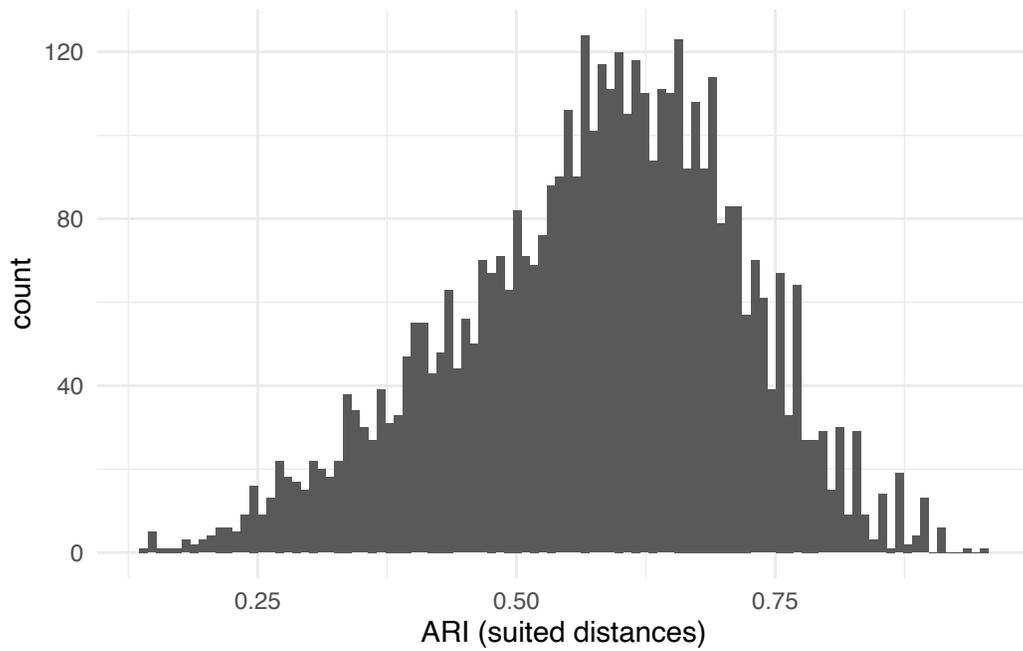

Figure S 12: Distribution of ARI values when results are filtered by suited distances

### 5.1.2 General model: effect of thematic foregrounding

What is the effect of thematic foregrounding for different feature types? For this model data was filtered by removing chunked WGCNA results and selecting distances with the highest average.

We fit a multilevel model with interaction between method ($Feature$) and the level of thematic foregrounding ($Level$), pooled by individual samples. In R library `brms` formula, it is `ari ~ 1 + Feature * Level + (1|sample)`. We use regularizing priors for 'intercept' and 'slope' coefficients as seen in the expanded model notation below. (We use dummy coding with `brms` interface for categorical variables, so $\beta$ coefficients represent *difference* between a $Feature$, $Level$ combination and reference 'intercept' which is doc2vec at level 1. $\beta = Normal(0, 0.1)$ does not expect any difference on average). We use $Level$ as a shorthand for the *level of thematic foregrounding* in notation. All further models have the same structure and priors.

$$ARI \sim Normal(\mu_i, \sigma_i)$$
$$\mu_i = \alpha + (\beta_f * Feature_i + \beta_{fl} * Level_i) + \beta_l * Level_i + \delta_{sample[i]}$$
$$\alpha = Normal(0.5, 0.1)$$
$$\beta_{f,l,fl} = Normal(0, 0.1)$$
$$\delta = Normal(0, \sigma_\delta)$$
$$\sigma_i = Exponential(1)$$
$$\sigma_\delta = Exponential(1)$$

We model the $Level$ of thematic foregrounding as categorical variable, and not ordinal, because we constructed the 'levels' artificially: there might not be any *order* in relationship between these levels. That said, modeling $Level$ via monotonic effects would still work and resulting models will be similar (as shown by leave-one-out cross validation in Table S6). Additionally, including varying slopes for individual samples does not improve model prediction much. It suggests that, across 100 samples of texts, methods and thematic foregrounding behaved similarly relative to each other. Since adding slopes to random effects can complicate fitting models and chain convergence, we instead only fit models grouped by samples.

Multilevel models with group-level effects for individual samples are always a better fit than those without. They allow to be more uncertain about the mean estimates, since clustering results notably differ from sample to sample.

Left-hand side of the Figure S13 shows posterior ARI means in each $Level$ and each $Feature$ type. Right-hand side shows the same relationship, but now the mean is taken marginal of samples: credible intervals are now much wider.

At the medium and strong thematic foregrounding three out of four feature types seem to behave similarly with doc2vec having an upper hand. We can directly compare their posterior distributions (Figure S14). Dotted lines represent the mean of distribution for each feature.

Table S 6: Leave-one-out results for group 1 models (feature*level)

|  | elpd_diff | se_diff |
|---|---|---|
| ari ~ level * type + (level + type \| sample_n) | 0.00 | 0.00 |
| ari ~ 1 + level * type + (1 + level + type + level:type \| sample_n) | -3.64 | 2.04 |
| ari ~ mo(level) * type + (1 \| sample_n) | -15.53 | 7.05 |
| ari ~ 1 + level * type + (1 \| sample_n) | -15.63 | 6.89 |
| ari ~ mo(level) * type | -213.19 | 21.42 |
| ari ~ 1 + level * type | -213.65 | 21.37 |
| ari ~ 1 + level + type | -514.73 | 31.39 |

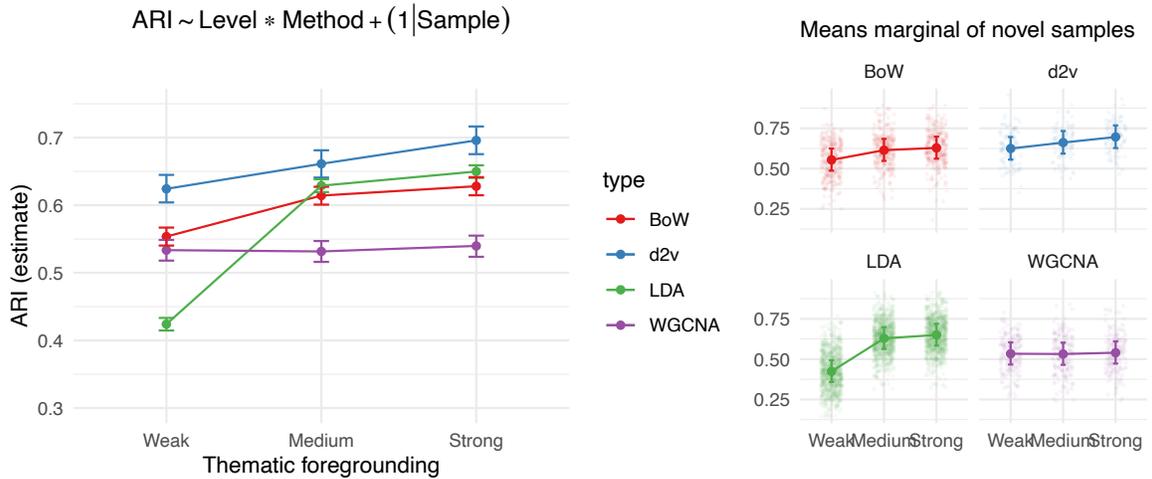

Figure S 13: Posterior predictions for methods behavior across thematic foregrounding levels (left). Pooled means, marginal of 100 sample runs (right).

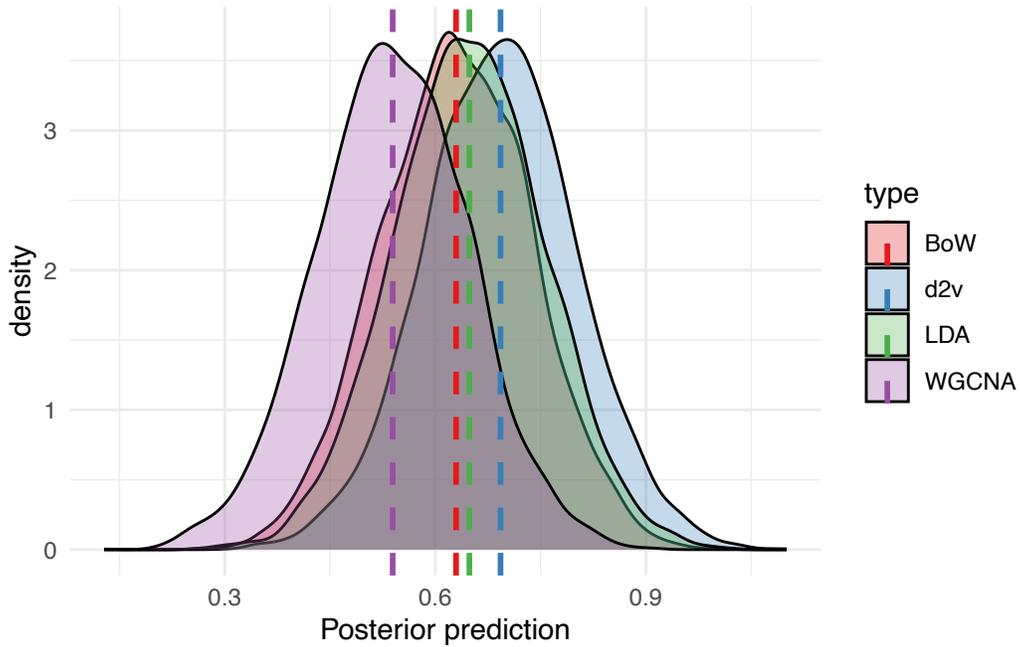

Figure S 14: Posterior distribution of ARI at strong thematic foregrounding for different features.

Sampling introduces considerable variation to the behavior of all features types. We can use posterior predictions to check differences in specific samples (10 samples drawn at random, Figure S15). Note that doc2vec has only one observation per sample for each level, but model uses grand mean to keep estimations conservative.

### 5.1.3 Overall best performance, distances filtered

To get an overall picture of comparable method performance we filter results by selected distances only. Figure S16 shows ARI boxplots per each of 51 combinations.

### 5.1.4 All combinations

Table S7 records all 291 combinations, without any filters, arranged by descending median ARI.

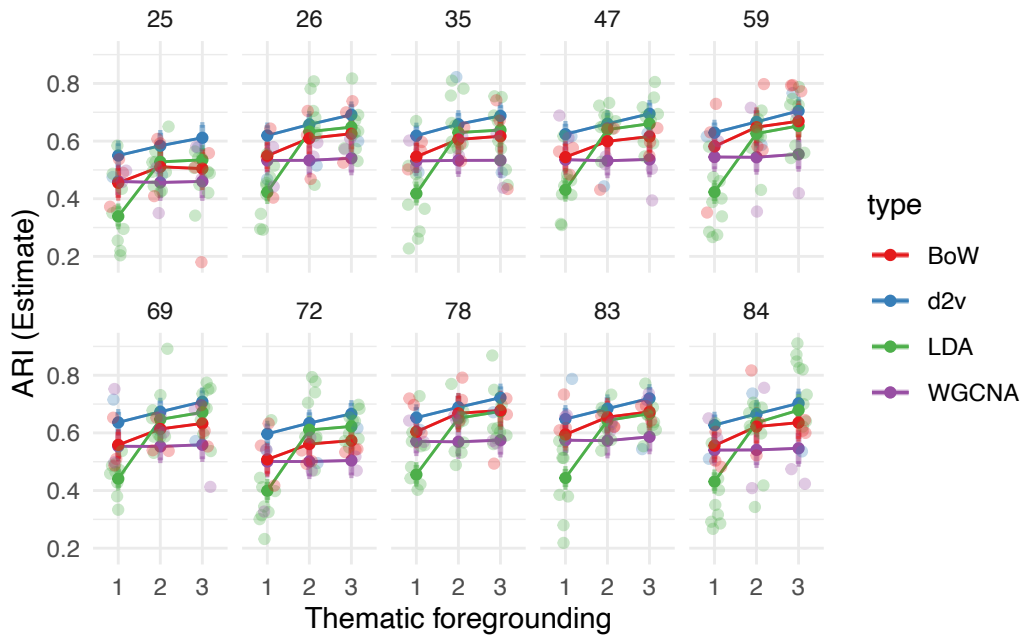

Figure S 15: Posterior predictions for 10 random individual samples, superimposed on the empirical data points. Varying slopes model.

Table S 7: Median performance of all 291 method + distance combinations

| rank | method | TF | distance | median.ARI | SD |
|------|--------|-----|----------|-----------|-----|
| 1 | d2v_lvl_3 | Strong | cosine | 0.703 | 0.092 |
| 2 | d2v_lvl_3 | Strong | euclidean | 0.687 | 0.097 |
| 3 | d2v_lvl_3 | Strong | cosine delta | 0.681 | 0.104 |
| 4 | LDA_lvl_3_50_5k | Strong | jensen-shannon | 0.677 | 0.107 |
| 5 | LDA_lvl_3_50_5k | Strong | delta | 0.672 | 0.099 |
| 6 | LDA_lvl_2_100_5k | Medium | manhattan | 0.671 | 0.096 |
| 7 | LDA_lvl_3_100_1k | Strong | jensen-shannon | 0.67 | 0.086 |
| 8 | d2v_lvl_2 | Medium | cosine | 0.668 | 0.093 |
| 9 | lvl_3_BoW | Strong | jensen-shannon | 0.665 | 0.11 |
| 10 | LDA_lvl_2_50_5k | Medium | jensen-shannon | 0.661 | 0.092 |
| 11 | d2v_lvl_2 | Medium | cosine delta | 0.657 | 0.125 |
| 12 | LDA_lvl_3_20_1k0 | Strong | jensen-shannon | 0.657 | 0.114 |
| 13 | lvl_3_BoW | Strong | jensen-shannon | 0.657 | 0.093 |
| 14 | LDA_lvl_3_100_5k | Strong | jensen-shannon | 0.656 | 0.074 |
| 15 | LDA_lvl_3_20_5k | Strong | jensen-shannon | 0.656 | 0.116 |
| 16 | LDA_lvl_3_50_1k | Strong | jensen-shannon | 0.656 | 0.099 |

| rank | method | TF | distance | median.ARI | SD |
|---|---|---|---|---|---|
| 17 | d2v_lvl_3 | Strong | delta | 0.655 | 0.089 |
| 18 | LDA_lvl_3_50_1k0 | Strong | delta | 0.654 | 0.098 |
| 19 | LDA_lvl_3_50_1k0 | Strong | jensen-shannon | 0.654 | 0.111 |
| 20 | lvl_2_BoW | Medium | jensen-shannon | 0.654 | 0.104 |
| 21 | d2v_lvl_3 | Strong | manhattan | 0.653 | 0.099 |
| 22 | LDA_lvl_2_100_5k | Medium | delta | 0.653 | 0.094 |
| 23 | d2v_lvl_1 | Weak | euclidean | 0.652 | 0.104 |
| 24 | d2v_lvl_2 | Medium | euclidean | 0.651 | 0.104 |
| 25 | LDA_lvl_3_100_1k | Strong | delta | 0.649 | 0.088 |
| 26 | LDA_lvl_3_100_1k | Strong | manhattan | 0.649 | 0.084 |
| 27 | LDA_lvl_2_50_1k0 | Medium | jensen-shannon | 0.648 | 0.1 |
| 28 | LDA_lvl_2_50_1k0 | Medium | delta | 0.643 | 0.099 |
| 29 | LDA_lvl_3_100_5k | Strong | manhattan | 0.643 | 0.095 |
| 30 | LDA_lvl_3_20_1k0 | Strong | delta | 0.642 | 0.103 |
| 31 | LDA_lvl_2_20_1k0 | Medium | delta | 0.639 | 0.096 |
| 32 | LDA_lvl_3_100_5k | Strong | delta | 0.639 | 0.085 |
| 33 | LDA_lvl_3_50_5k | Strong | manhattan | 0.637 | 0.115 |
| 34 | lvl_2_BoW | Medium | jensen-shannon | 0.637 | 0.098 |
| 35 | d2v_lvl_2 | Medium | manhattan | 0.636 | 0.099 |
| 36 | LDA_lvl_2_20_1k0 | Medium | jensen-shannon | 0.636 | 0.111 |
| 37 | LDA_lvl_2_20_5k | Medium | jensen-shannon | 0.636 | 0.106 |
| 38 | LDA_lvl_2_20_1k0 | Medium | manhattan | 0.635 | 0.129 |
| 39 | LDA_lvl_2_50_5k | Medium | delta | 0.635 | 0.093 |
| 40 | LDA_lvl_2_100_1k | Medium | jensen-shannon | 0.634 | 0.09 |
| 41 | lvl_1_BoW | Weak | cosine delta | 0.634 | 0.094 |
| 42 | LDA_lvl_3_20_5k | Strong | manhattan | 0.632 | 0.104 |
| 43 | d2v_lvl_2 | Medium | delta | 0.631 | 0.105 |
| 44 | lvl_1_BoW | Weak | cosine delta | 0.631 | 0.095 |
| 45 | LDA_lvl_3_20_1k | Strong | jensen-shannon | 0.628 | 0.106 |
| 46 | LDA_lvl_3_20_1k0 | Strong | manhattan | 0.628 | 0.107 |
| 47 | d2v_lvl_1 | Weak | cosine | 0.627 | 0.108 |
| 48 | LDA_lvl_2_100_1k | Medium | manhattan | 0.627 | 0.09 |
| 49 | d2v_lvl_1 | Weak | cosine delta | 0.626 | 0.105 |
| 50 | d2v_lvl_1 | Weak | manhattan | 0.626 | 0.108 |
| 51 | lvl_2_BoW | Medium | cosine delta | 0.625 | 0.104 |
| 52 | LDA_lvl_2_100_1k0 | Medium | delta | 0.624 | 0.112 |
| 53 | LDA_lvl_2_50_1k | Medium | jensen-shannon | 0.623 | 0.094 |
| 54 | LDA_lvl_3_100_1k0 | Strong | delta | 0.623 | 0.096 |
| 55 | lvl_1_BoW | Weak | jensen-shannon | 0.623 | 0.092 |
| 56 | d2v_lvl_1 | Weak | delta | 0.622 | 0.109 |
| 57 | LDA_lvl_2_20_5k | Medium | manhattan | 0.622 | 0.112 |

| rank | method | TF | distance | median.ARI | SD |
|---|---|---|---|---|---|
| 58 | LDA_lvl_2_20_1k | Medium | jensen-shannon | 0.621 | 0.113 |
| 59 | LDA_lvl_2_100_1k0 | Medium | jensen-shannon | 0.615 | 0.104 |
| 60 | LDA_lvl_3_20_5k | Strong | delta | 0.615 | 0.092 |
| 61 | lvl_3_BoW | Strong | cosine delta | 0.615 | 0.108 |
| 62 | LDA_lvl_2_100_5k | Medium | jensen-shannon | 0.614 | 0.08 |
| 63 | lvl_2_BoW | Medium | cosine delta | 0.614 | 0.113 |
| 64 | LDA_lvl_3_50_1k | Strong | manhattan | 0.609 | 0.109 |
| 65 | LDA_lvl_3_20_1k | Strong | manhattan | 0.605 | 0.116 |
| 66 | LDA_lvl_2_50_5k | Medium | manhattan | 0.604 | 0.095 |
| 67 | LDA_lvl_3_50_1k | Strong | delta | 0.604 | 0.095 |
| 68 | lvl_1_BoW | Weak | jensen-shannon | 0.601 | 0.09 |
| 69 | LDA_lvl_2_20_5k | Medium | delta | 0.6 | 0.09 |
| 70 | lvl_3_BoW | Strong | cosine delta | 0.599 | 0.087 |
| 71 | WGCNA_lvl2_5k_ | Medium | cosine delta | 0.599 | 0.098 |
| 72 | WGCNA_lvl2_5k_ | Medium | cosine | 0.599 | 0.098 |
| 73 | lvl_1_BoW | Weak | manhattan | 0.596 | 0.09 |
| 74 | LDA_lvl_2_100_1k | Medium | delta | 0.595 | 0.09 |
| 75 | LDA_lvl_3_100_1k0 | Strong | jensen-shannon | 0.592 | 0.099 |
| 76 | LDA_lvl_3_20_1k | Strong | delta | 0.592 | 0.089 |
| 77 | lvl_1_BoW | Weak | delta | 0.592 | 0.096 |
| 78 | LDA_lvl_2_50_1k0 | Medium | manhattan | 0.59 | 0.105 |
| 79 | LDA_lvl_2_100_5k | Medium | cosine delta | 0.588 | 0.102 |
| 80 | lvl_1_BoW | Weak | delta | 0.588 | 0.123 |
| 81 | LDA_lvl_3_100_5k | Strong | cosine delta | 0.585 | 0.091 |
| 82 | WGCNA_lvl1_5k_ | Weak | delta | 0.583 | 0.086 |
| 83 | WGCNA_lvl1_5k_ | Weak | manhattan | 0.583 | 0.086 |
| 84 | WGCNA_lvl3_5k_ | Strong | cosine delta | 0.583 | 0.101 |
| 85 | WGCNA_lvl3_5k_ | Strong | cosine | 0.583 | 0.101 |
| 86 | LDA_lvl_2_20_1k | Medium | manhattan | 0.582 | 0.125 |
| 87 | lvl_1_BoW | Weak | manhattan | 0.582 | 0.093 |
| 88 | WGCNA_lvl1_5k_ | Weak | cosine delta | 0.581 | 0.104 |
| 89 | WGCNA_lvl1_5k_ | Weak | cosine | 0.581 | 0.104 |
| 90 | lvl_3_BoW | Strong | delta | 0.579 | 0.116 |
| 91 | LDA_lvl_3_50_1k0 | Strong | manhattan | 0.577 | 0.117 |
| 92 | LDA_lvl_3_100_1k | Strong | cosine delta | 0.576 | 0.076 |
| 93 | lvl_3_BoW | Strong | jensen-shannon | 0.575 | 0.101 |
| 94 | lvl_2_BoW | Medium | delta | 0.572 | 0.121 |
| 95 | LDA_lvl_3_100_1k0 | Strong | cosine delta | 0.571 | 0.094 |
| 96 | LDA_lvl_2_50_1k | Medium | delta | 0.566 | 0.107 |
| 97 | lvl_2_BoW | Medium | jensen-shannon | 0.564 | 0.091 |
| 98 | lvl_2_BoW | Medium | manhattan | 0.564 | 0.117 |

| rank | method | TF | distance | median.ARI | SD |
|------|--------|-----|----------|-----------|-----|
| 99 | LDA_lvl_3_20_1k0 | Strong | cosine delta | 0.563 | 0.095 |
| 100 | LDA_lvl_3_20_5k | Strong | cosine delta | 0.56 | 0.088 |
| 101 | LDA_lvl_2_20_1k | Medium | delta | 0.558 | 0.101 |
| 102 | lvl_1_BoW | Weak | cosine delta | 0.558 | 0.092 |
| 103 | LDA_lvl_3_20_1k0 | Strong | cosine | 0.556 | 0.121 |
| 104 | lvl_3_BoW | Strong | manhattan | 0.556 | 0.124 |
| 105 | LDA_lvl_2_20_1k0 | Medium | cosine delta | 0.555 | 0.098 |
| 106 | lvl_3_BoW | Strong | manhattan | 0.554 | 0.12 |
| 107 | WGCNA_lvl2_5k_ | Medium | delta | 0.552 | 0.109 |
| 108 | WGCNA_lvl2_5k_ | Medium | manhattan | 0.552 | 0.109 |
| 109 | LDA_lvl_1_100_1k | Weak | delta | 0.551 | 0.088 |
| 110 | LDA_lvl_2_50_1k | Medium | manhattan | 0.551 | 0.096 |
| 111 | LDA_lvl_2_100_1k0 | Medium | cosine delta | 0.548 | 0.103 |
| 112 | LDA_lvl_3_20_1k | Strong | cosine | 0.546 | 0.128 |
| 113 | LDA_lvl_1_100_5k | Weak | delta | 0.545 | 0.085 |
| 114 | LDA_lvl_2_100_1k | Medium | cosine delta | 0.544 | 0.079 |
| 115 | lvl_1_BoW | Weak | delta | 0.543 | 0.099 |
| 116 | WGCNA_lvl2_5k_ | Medium | euclidean | 0.543 | 0.107 |
| 117 | LDA_lvl_1_100_1k0 | Weak | delta | 0.542 | 0.096 |
| 118 | LDA_lvl_3_20_5k | Strong | cosine | 0.541 | 0.126 |
| 119 | LDA_lvl_2_20_5k | Medium | cosine delta | 0.539 | 0.106 |
| 120 | WGCNA_lvl3_5k_ | Strong | delta | 0.538 | 0.102 |
| 121 | WGCNA_lvl3_5k_ | Strong | manhattan | 0.538 | 0.102 |
| 122 | LDA_lvl_3_50_5k | Strong | cosine delta | 0.53 | 0.102 |
| 123 | LDA_lvl_1_100_1k | Weak | manhattan | 0.529 | 0.087 |
| 124 | lvl_2_BoW | Medium | cosine delta | 0.529 | 0.092 |
| 125 | lvl_2_BoW | Medium | manhattan | 0.529 | 0.107 |
| 126 | WGCNA_lvl3_5k_ | Strong | euclidean | 0.529 | 0.124 |
| 127 | LDA_lvl_3_50_1k | Strong | cosine delta | 0.527 | 0.082 |
| 128 | lvl_2_BoW | Medium | delta | 0.525 | 0.134 |
| 129 | lvl_3_BoW | Strong | cosine delta | 0.522 | 0.106 |
| 130 | LDA_lvl_2_20_1k0 | Medium | cosine | 0.519 | 0.132 |
| 131 | LDA_lvl_3_100_1k0 | Strong | manhattan | 0.519 | 0.108 |
| 132 | LDA_lvl_3_50_1k0 | Strong | cosine delta | 0.518 | 0.092 |
| 133 | LDA_lvl_1_100_1k | Weak | jensen-shannon | 0.514 | 0.094 |
| 134 | LDA_lvl_3_20_1k | Strong | cosine delta | 0.511 | 0.09 |
| 135 | LDA_lvl_2_20_5k | Medium | cosine | 0.505 | 0.117 |
| 136 | lvl_2_BoW | Medium | delta | 0.505 | 0.097 |
| 137 | LDA_lvl_1_100_1k | Weak | cosine delta | 0.503 | 0.088 |
| 138 | LDA_lvl_2_20_1k | Medium | cosine delta | 0.501 | 0.083 |
| 139 | LDA_lvl_2_50_1k | Medium | cosine delta | 0.501 | 0.095 |

| rank | method | TF | distance | median.ARI | SD |
|------|--------|------|----------|------------|------|
| 140 | WGCNA_lvl1_1k_ | Weak | cosine delta | 0.498 | 0.104 |
| 141 | WGCNA_lvl1_1k_ | Weak | cosine | 0.498 | 0.104 |
| 142 | WGCNA_lvl1_5k_ | Weak | euclidean | 0.497 | 0.147 |
| 143 | LDA_lvl_1_100_5k | Weak | jensen-shannon | 0.495 | 0.109 |
| 144 | lvl_3_BoW | Strong | delta | 0.495 | 0.106 |
| 145 | LDA_lvl_2_50_1k0 | Medium | cosine delta | 0.494 | 0.117 |
| 146 | lvl_3_BoW | Strong | manhattan | 0.49 | 0.103 |
| 147 | LDA_lvl_2_50_5k | Medium | cosine delta | 0.489 | 0.09 |
| 148 | LDA_lvl_1_50_1k | Weak | delta | 0.484 | 0.084 |
| 149 | LDA_lvl_2_100_1k0 | Medium | manhattan | 0.482 | 0.131 |
| 150 | LDA_lvl_1_100_5k | Weak | manhattan | 0.481 | 0.101 |
| 151 | LDA_lvl_2_20_1k | Medium | cosine | 0.479 | 0.114 |
| 152 | lvl_1_BoW | Weak | manhattan | 0.475 | 0.096 |
| 153 | WGCNA_lvl3_1k_ | Strong | cosine delta | 0.473 | 0.11 |
| 154 | WGCNA_lvl3_1k_ | Strong | cosine | 0.473 | 0.11 |
| 155 | LDA_lvl_1_100_5k | Weak | cosine delta | 0.471 | 0.09 |
| 156 | lvl_3_BoW | Strong | delta | 0.469 | 0.137 |
| 157 | LDA_lvl_3_100_1k | Strong | cosine | 0.467 | 0.118 |
| 158 | WGCNA_lvl2_1k_ | Medium | cosine delta | 0.467 | 0.102 |
| 159 | WGCNA_lvl2_1k_ | Medium | cosine | 0.467 | 0.102 |
| 160 | LDA_lvl_3_20_5k | Strong | euclidean | 0.464 | 0.134 |
| 161 | LDA_lvl_1_100_1k0 | Weak | jensen-shannon | 0.462 | 0.106 |
| 162 | WGCNA_lvl1_1k_ | Weak | euclidean | 0.462 | 0.115 |
| 163 | LDA_lvl_3_50_1k | Strong | cosine | 0.461 | 0.114 |
| 164 | LDA_lvl_3_20_1k | Strong | euclidean | 0.46 | 0.136 |
| 165 | lvl_1_BoW | Weak | jensen-shannon | 0.458 | 0.085 |
| 166 | LDA_lvl_2_50_1k | Medium | cosine | 0.453 | 0.095 |
| 167 | WGCNA_lvl1_1k_ | Weak | delta | 0.453 | 0.094 |
| 168 | WGCNA_lvl1_1k_ | Weak | manhattan | 0.453 | 0.094 |
| 169 | LDA_lvl_2_20_5k | Medium | euclidean | 0.451 | 0.133 |
| 170 | lvl_2_BoW | Medium | manhattan | 0.451 | 0.1 |
| 171 | LDA_lvl_1_100_1k0 | Weak | cosine delta | 0.449 | 0.09 |
| 172 | LDA_lvl_1_50_5k | Weak | jensen-shannon | 0.442 | 0.1 |
| 173 | LDA_lvl_1_50_5k | Weak | delta | 0.439 | 0.095 |
| 174 | LDA_lvl_3_20_1k0 | Strong | euclidean | 0.433 | 0.149 |
| 175 | LDA_lvl_2_50_5k | Medium | cosine | 0.429 | 0.099 |
| 176 | WGCNA_lvl2_1k_ | Medium | delta | 0.429 | 0.109 |
| 177 | WGCNA_lvl2_1k_ | Medium | manhattan | 0.429 | 0.109 |
| 178 | WGCNA_lvl3_1k_ | Strong | delta | 0.429 | 0.101 |
| 179 | WGCNA_lvl3_1k_ | Strong | manhattan | 0.429 | 0.101 |
| 180 | WGCNA_lvl3_1k_ | Strong | euclidean | 0.428 | 0.105 |

| rank | method | TF | distance | median.ARI | SD |
|---|---|---|---|---|---|
| 181 | LDA_lvl_2_20_1k0 | Medium | euclidean | 0.425 | 0.158 |
| 182 | LDA_lvl_1_100_1k0 | Weak | manhattan | 0.419 | 0.092 |
| 183 | LDA_lvl_1_50_1k | Weak | manhattan | 0.417 | 0.101 |
| 184 | LDA_lvl_2_100_1k | Medium | cosine | 0.416 | 0.123 |
| 185 | LDA_lvl_3_50_5k | Strong | cosine | 0.411 | 0.103 |
| 186 | LDA_lvl_1_50_1k | Weak | jensen-shannon | 0.41 | 0.106 |
| 187 | WGCNA_lvl2_1k_ | Medium | euclidean | 0.409 | 0.106 |
| 188 | LDA_lvl_1_50_5k | Weak | manhattan | 0.399 | 0.112 |
| 189 | LDA_lvl_1_20_5k | Weak | jensen-shannon | 0.398 | 0.088 |
| 190 | LDA_lvl_2_20_1k | Medium | euclidean | 0.397 | 0.119 |
| 191 | LDA_lvl_1_50_1k0 | Weak | jensen-shannon | 0.393 | 0.101 |
| 192 | LDA_lvl_1_20_1k | Weak | delta | 0.39 | 0.102 |
| 193 | LDA_lvl_1_50_1k0 | Weak | delta | 0.386 | 0.094 |
| 194 | LDA_lvl_1_20_5k | Weak | delta | 0.374 | 0.092 |
| 195 | LDA_lvl_3_50_5k | Strong | cosine | 0.371 | 0.113 |
| 196 | LDA_lvl_1_20_1k | Weak | jensen-shannon | 0.366 | 0.097 |
| 197 | LDA_lvl_2_50_1k0 | Medium | cosine | 0.364 | 0.105 |
| 198 | LDA_lvl_1_20_1k | Weak | cosine delta | 0.363 | 0.087 |
| 199 | LDA_lvl_1_50_1k0 | Weak | manhattan | 0.363 | 0.104 |
| 200 | LDA_lvl_2_100_5k | Medium | cosine | 0.356 | 0.087 |
| 201 | LDA_lvl_1_20_1k | Weak | manhattan | 0.355 | 0.113 |
| 202 | LDA_lvl_1_20_5k | Weak | manhattan | 0.351 | 0.102 |
| 203 | LDA_lvl_3_100_5k | Strong | cosine | 0.348 | 0.104 |
| 204 | LDA_lvl_1_50_1k | Weak | cosine delta | 0.338 | 0.087 |
| 205 | LDA_lvl_1_20_1k | Weak | cosine | 0.33 | 0.081 |
| 206 | LDA_lvl_1_20_5k | Weak | cosine | 0.328 | 0.1 |
| 207 | LDA_lvl_1_20_1k0 | Weak | delta | 0.325 | 0.096 |
| 208 | LDA_lvl_1_20_1k0 | Weak | jensen-shannon | 0.308 | 0.098 |
| 209 | LDA_lvl_3_100_1k0 | Strong | cosine | 0.291 | 0.084 |
| 210 | lvl_1_BoW | Weak | cosine | 0.291 | 0.065 |
| 211 | LDA_lvl_1_50_1k | Weak | cosine | 0.289 | 0.069 |
| 212 | lvl_1_BoW | Weak | cosine | 0.289 | 0.063 |
| 213 | LDA_lvl_1_20_1k0 | Weak | manhattan | 0.286 | 0.093 |
| 214 | LDA_lvl_1_20_1k0 | Weak | cosine | 0.276 | 0.091 |
| 215 | LDA_lvl_1_50_5k | Weak | cosine delta | 0.275 | 0.094 |
| 216 | LDA_lvl_1_20_5k | Weak | cosine delta | 0.27 | 0.077 |
| 217 | LDA_lvl_2_100_1k0 | Medium | cosine | 0.27 | 0.094 |
| 218 | LDA_lvl_1_20_1k | Weak | euclidean | 0.266 | 0.085 |
| 219 | LDA_lvl_1_50_1k | Weak | cosine delta | 0.251 | 0.082 |
| 220 | LDA_lvl_1_50_1k0 | Weak | cosine | 0.245 | 0.095 |
| 221 | LDA_lvl_1_100_1k0 | Weak | cosine | 0.244 | 0.096 |

| rank | method | TF | distance | median.ARI | SD |
|------|--------|-----|----------|-----------|-----|
| 222 | LDA_lvl_1_100_1k | Weak | cosine | 0.243 | 0.066 |
| 223 | LDA_lvl_1_50_5k | Weak | cosine | 0.243 | 0.096 |
| 224 | WGCNA_lvl1_5k_chunked | Weak | cosine delta | 0.242 | 0.029 |
| 225 | lvl_1_BoW | Weak | cosine | 0.238 | 0.055 |
| 226 | LDA_lvl_1_100_5k | Weak | cosine | 0.236 | 0.078 |
| 227 | LDA_lvl_1_20_1k0 | Weak | cosine delta | 0.232 | 0.078 |
| 228 | LDA_lvl_3_50_5k | Strong | euclidean | 0.232 | 0.138 |
| 229 | WGCNA_lvl1_1k_chunked | Weak | cosine delta | 0.23 | 0.029 |
| 230 | WGCNA_lvl1_5k_chunked | Weak | cosine | 0.23 | 0.027 |
| 231 | WGCNA_lvl2_5k_chunked | Medium | cosine delta | 0.23 | 0.02 |
| 232 | WGCNA_lvl3_5k_chunked | Strong | cosine delta | 0.229 | 0.021 |
| 233 | LDA_lvl_1_50_1k | Weak | euclidean | 0.226 | 0.09 |
| 234 | lvl_1_BoW | Weak | euclidean | 0.226 | 0.064 |
| 235 | WGCNA_lvl1_5k_chunked | Weak | manhattan | 0.222 | 0.046 |
| 236 | lvl_1_BoW | Weak | euclidean | 0.221 | 0.065 |
| 237 | WGCNA_lvl1_5k_chunked | Weak | euclidean | 0.221 | 0.056 |
| 238 | WGCNA_lvl3_1k_chunked | Strong | cosine delta | 0.218 | 0.025 |
| 239 | lvl_1_BoW | Weak | euclidean | 0.217 | 0.063 |
| 240 | LDA_lvl_3_50_1k | Strong | euclidean | 0.216 | 0.139 |
| 241 | lvl_3_BoW | Strong | cosine | 0.215 | 0.083 |
| 242 | WGCNA_lvl2_1k_chunked | Medium | cosine delta | 0.207 | 0.024 |
| 243 | lvl_3_BoW | Strong | cosine | 0.206 | 0.086 |
| 244 | WGCNA_lvl1_5k_chunked | Weak | delta | 0.206 | 0.109 |
| 245 | WGCNA_lvl2_5k_chunked | Medium | delta | 0.204 | 0.032 |
| 246 | WGCNA_lvl2_5k_chunked | Medium | euclidean | 0.204 | 0.032 |
| 247 | WGCNA_lvl2_5k_chunked | Medium | manhattan | 0.204 | 0.032 |
| 248 | LDA_lvl_1_20_5k | Weak | euclidean | 0.203 | 0.088 |
| 249 | WGCNA_lvl2_5k_chunked | Medium | cosine | 0.203 | 0.025 |
| 250 | WGCNA_lvl3_5k_chunked | Strong | cosine | 0.202 | 0.024 |
| 251 | WGCNA_lvl1_1k_chunked | Weak | cosine | 0.198 | 0.028 |
| 252 | lvl_3_BoW | Strong | cosine | 0.197 | 0.067 |
| 253 | LDA_lvl_2_50_1k0 | Medium | euclidean | 0.196 | 0.129 |
| 254 | WGCNA_lvl3_5k_chunked | Strong | delta | 0.196 | 0.036 |
| 255 | WGCNA_lvl3_5k_chunked | Strong | euclidean | 0.196 | 0.036 |
| 256 | WGCNA_lvl3_5k_chunked | Strong | manhattan | 0.196 | 0.036 |
| 257 | WGCNA_lvl1_1k_chunked | Weak | delta | 0.191 | 0.043 |
| 258 | WGCNA_lvl1_1k_chunked | Weak | euclidean | 0.191 | 0.043 |
| 259 | WGCNA_lvl1_1k_chunked | Weak | manhattan | 0.191 | 0.043 |
| 260 | LDA_lvl_2_50_5k | Medium | euclidean | 0.19 | 0.134 |
| 261 | WGCNA_lvl3_1k_chunked | Strong | cosine | 0.184 | 0.027 |
| 262 | LDA_lvl_3_50_1k0 | Strong | euclidean | 0.183 | 0.119 |

| rank | method | TF | distance | median.ARI | SD |
|---|---|---|---|---|---|
| 263 | LDA_lvl_2_50_1k | Medium | euclidean | 0.182 | 0.148 |
| 264 | WGCNA_lvl2_1k_chunked | Medium | cosine | 0.181 | 0.029 |
| 265 | lvl_2_BoW | Medium | cosine | 0.18 | 0.077 |
| 266 | lvl_3_BoW | Strong | euclidean | 0.177 | 0.07 |
| 267 | lvl_2_BoW | Medium | euclidean | 0.175 | 0.054 |
| 268 | lvl_2_BoW | Medium | cosine | 0.174 | 0.061 |
| 269 | WGCNA_lvl2_1k_chunked | Medium | delta | 0.174 | 0.037 |
| 270 | WGCNA_lvl2_1k_chunked | Medium | euclidean | 0.174 | 0.037 |
| 271 | WGCNA_lvl2_1k_chunked | Medium | manhattan | 0.174 | 0.037 |
| 272 | lvl_2_BoW | Medium | cosine | 0.173 | 0.074 |
| 273 | lvl_3_BoW | Strong | euclidean | 0.171 | 0.054 |
| 274 | lvl_3_BoW | Strong | euclidean | 0.162 | 0.058 |
| 275 | WGCNA_lvl3_1k_chunked | Strong | delta | 0.162 | 0.031 |
| 276 | WGCNA_lvl3_1k_chunked | Strong | euclidean | 0.162 | 0.031 |
| 277 | WGCNA_lvl3_1k_chunked | Strong | manhattan | 0.162 | 0.031 |
| 278 | lvl_2_BoW | Medium | euclidean | 0.154 | 0.05 |
| 279 | lvl_2_BoW | Medium | euclidean | 0.149 | 0.047 |
| 280 | LDA_lvl_1_20_1k0 | Weak | euclidean | 0.143 | 0.076 |
| 281 | LDA_lvl_1_100_1k0 | Weak | euclidean | 0.031 | 0.078 |
| 282 | LDA_lvl_1_50_5k | Weak | euclidean | 0.03 | 0.102 |
| 283 | LDA_lvl_1_50_1k0 | Weak | euclidean | 0.02 | 0.08 |
| 284 | LDA_lvl_2_100_1k0 | Medium | euclidean | 0.006 | 0.057 |
| 285 | LDA_lvl_1_100_5k | Weak | euclidean | 0.005 | 0.066 |
| 286 | LDA_lvl_3_100_1k0 | Strong | euclidean | 0.005 | 0.072 |
| 287 | LDA_lvl_3_100_5k | Strong | euclidean | 0.004 | 0.061 |
| 288 | LDA_lvl_2_100_1k | Medium | euclidean | 0.003 | 0.068 |
| 289 | LDA_lvl_2_100_5k | Medium | euclidean | 0.003 | 0.055 |
| 290 | LDA_lvl_3_100_1k | Strong | euclidean | 0.003 | 0.066 |
| 291 | LDA_lvl_1_100_1k | Weak | euclidean | 0.002 | 0.065 |

### 5.1.5 LDA

For the LDA model we want to know the effect of number of topics and MFWs used to prepare training DTM. We fit the following multilevel interaction model:

```
ARI ~ 1 + level*MFWs*topics + (1 + | sample)
```

First, we look at direct effects of *MFWs* and *topics* on ARI across all thematic foregrounding *levels* and marginal of novel samples (Figure S17). It appears that LDA with larger number of topics and smaller number of MFWs performs slightly better on average. LDA models with 100 topics also show the smallest variance in performance across sampling runs. These effects,

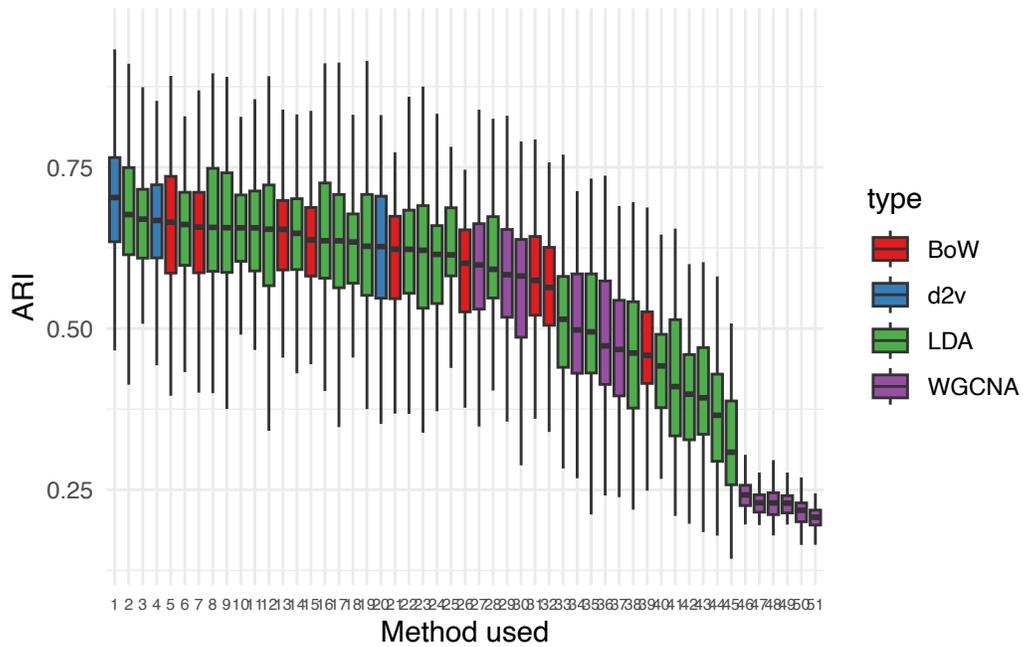

Figure S 16: Empirical distributions of method combinations. Error bars correspond to 95%
CI. Filtered distances.

Table S 8: Leave-one-out results for LDA models.

|  | elpd_diff | se_diff |
|---|---|---|
| ari ~ level * mfw_t * topics + (1 + level + mfw_t + topics \| sample_n) | 0.00 | 0.00 |
| ari ~ level * mfw_t * topics + (1 \| sample_n) | -24.75 | 8.19 |
| ari ~ level + mfw_t + topics + (1 + level + mfw_t + topics \| sample_n) | -150.45 | 17.62 |
| ari ~ level * mfw_t * topics | -173.71 | 20.14 |
| ari ~ level * mfw_t + topics | -280.31 | 24.17 |
| ari ~ level + mfw_t + topics | -287.66 | 24.51 |

however, mostly come from the corpus with weak thematic foregrounding as seen on Figure S18.

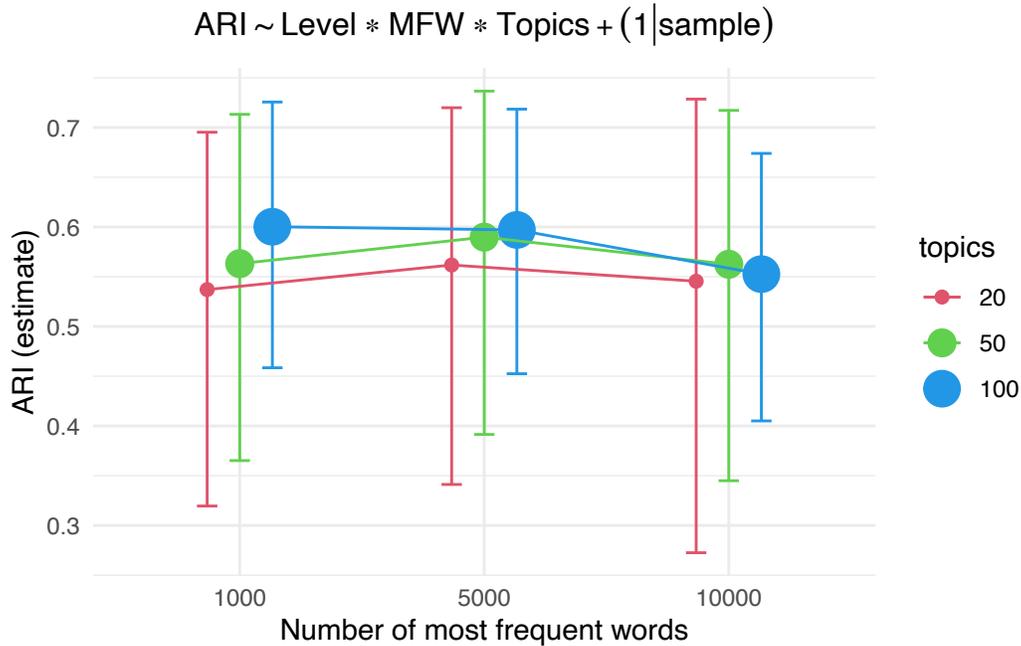

Figure S 17: Posterior predictions for LDA performance, marginal of samples.

Bars mark posterior .95 CI, shaded dots show empirical LDA results. We see that, again, the level of thematic foregrounding has the largest influence on LDA performance. At medium and strong levels, however, the impact of topics and MFWs is not clear. It seems that, on average, an increase in the number of topics for the small number of features tends to improve clustering, while the effect is reversed for large number of features (smaller number of topics have a slight edge, see Figure S19). Overall, the choice of number of topics and features is more critical in a corpus without pre-processing and becomes less influential when features are foregrounded.

### 5.1.6 Bags of words

Again, we fit a Bayesian multilevel interaction model. Two factors drive the performance of bag-of-words features: level of thematic foregrounding and the length of vector (number of MFWs). In `brms` model notation:

```
ari ~ level*MFWs + (1|sample_no)
```

Figure S20 shows that clustering with bag of words improve on average with longer vectors: since there is no algorithm that summarizes similarity in individual words behavior, word frequences are more dependent on diverse lexical pools and sparse DTMs. It might be a suboptimal way to model texts, since the final clustering would rely on groups of present/absent words rather than the actual distribution. Also we would expect results to plateau if the length of bag of words is increased further. The plateau is better visible when posterior estimates are taken as average of foregrounding levels and marginal of samples (Figure S21).

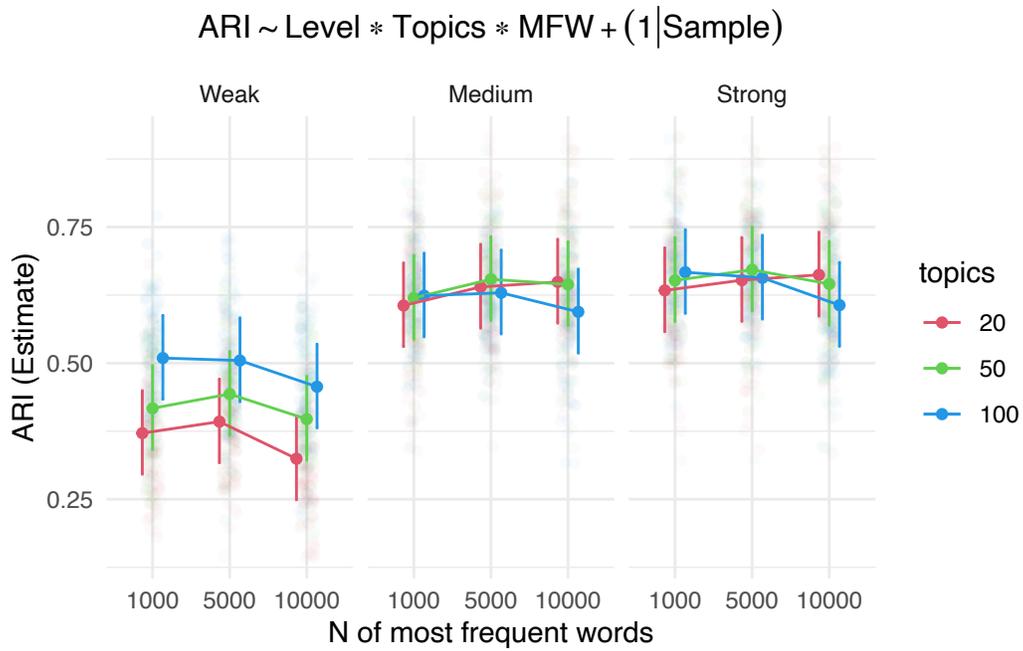

Figure S 18: Posterior predictions for LDA performance grouped by thematic foregrounding, marginal of samples.

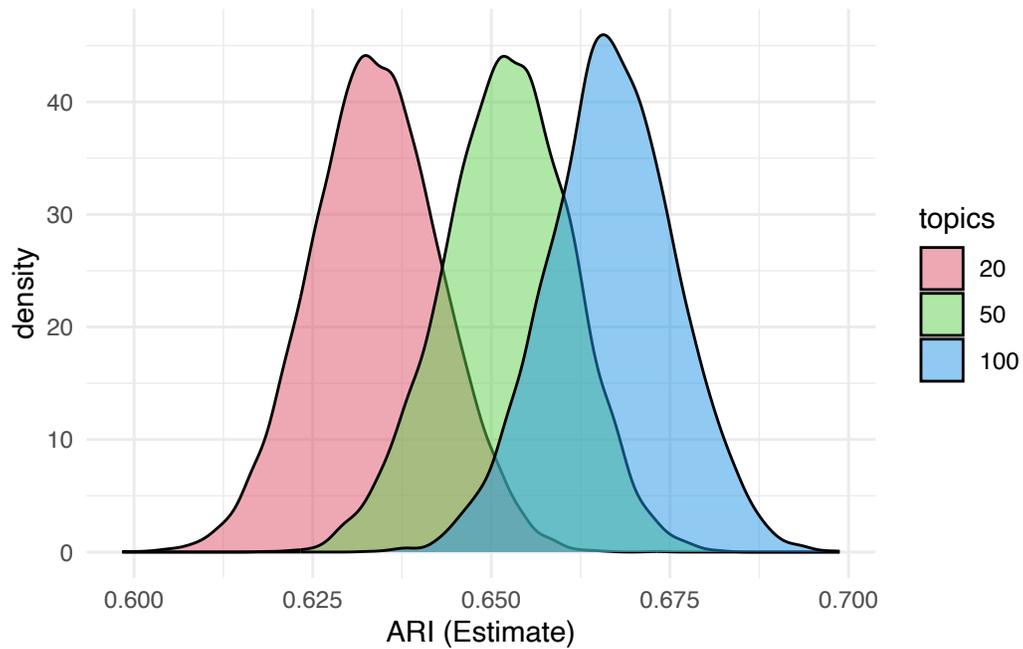

Figure S 19: LDA posterior means at strong thematic foregrounding and 1000 MFWs.

Table S 9: Leave-one-out results for BoW group of models.

| | elpd_diff | se_diff |
|---|---|---|
| ari ~ level * mfw_t + (1 + level + mfw_t \| sample_n) | 0.00 | 0.00 |
| ari ~ level + mfw_t + (1 + level + mfw_t \| sample_n) | -11.99 | 5.21 |
| ari ~ level * mfw_t + (1 \| sample_n) | -29.54 | 8.38 |
| ari ~ level * mfw_t | -142.47 | 17.73 |
| ari ~ level + mfw_t | -145.77 | 17.95 |

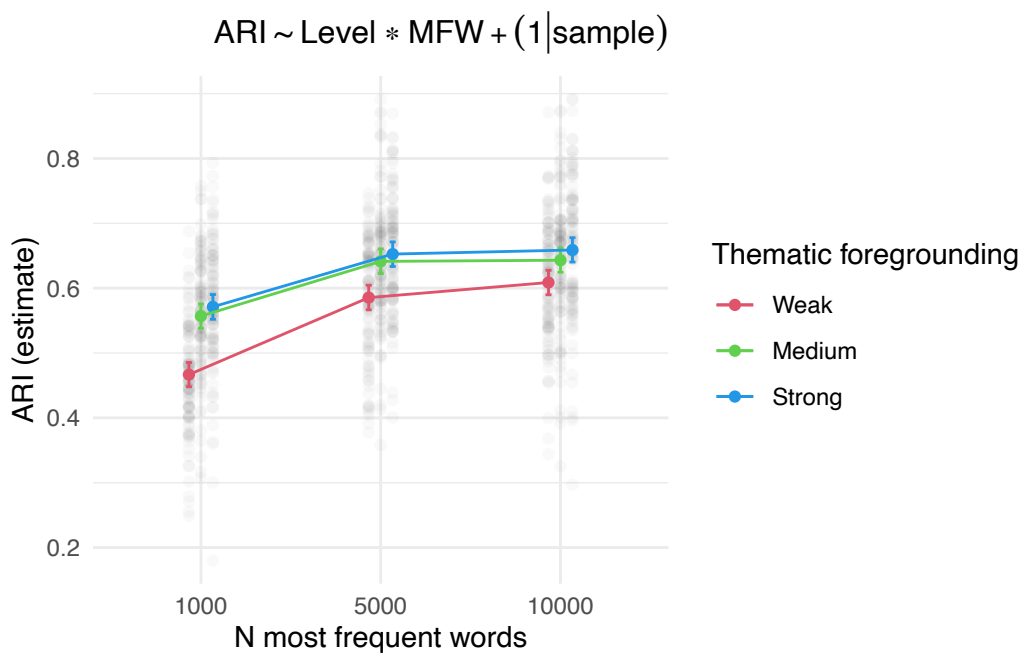

Figure S 20: Bag-of-words posterior predictions, superimposed on empirical observations.

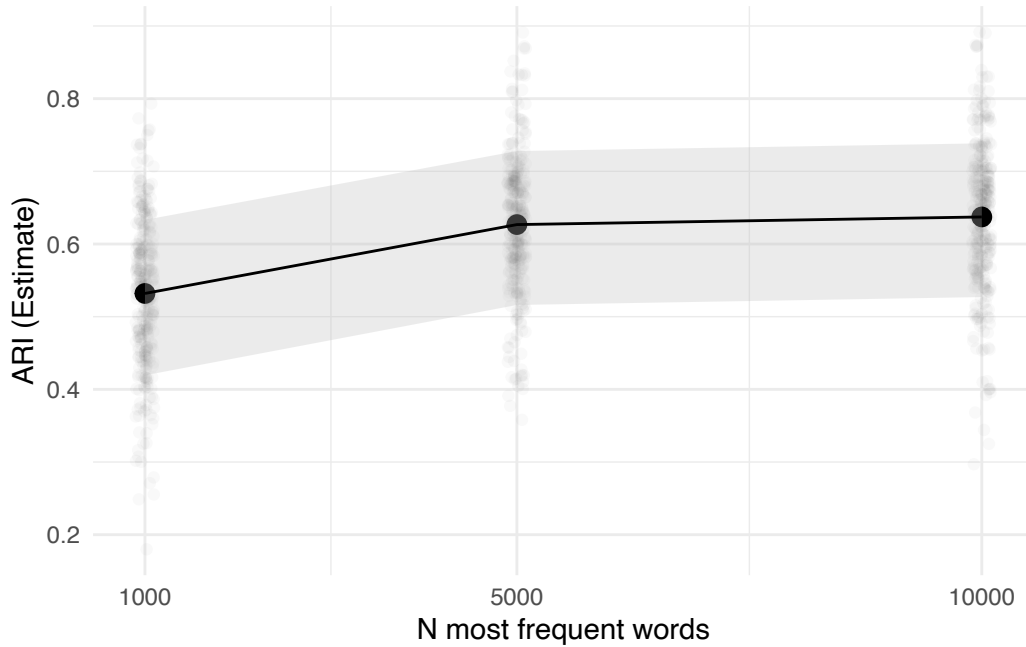

Figure S 21: Bag-of-words posterior means, summarized for the levels of thematic foregrounding and marginal of the samples of novels.

### 5.1.7 WGCNA

We model three factors in WGCNA performance: chunking, level and MFWs:

```
ari ~ chunking*level*MFWs + (1 + chunking + level + MFWs | sample_no)
```

First, Figure S22 clearly confirms that that chunking texts drastically reduces the performance of clustering with WGCNA modules, because of greedy module identification problem (see Section 2.3).

Figure S23 shows posterior means for different cut-offs of MFWs and thematic foregrounding levels (only for models without chunking).

Non-chunked WGCNA, on average, benefits mostly from medium thematic foregrounding and increasing MFWs.

Table S 10: Leave-one-out results for WGCNA group of models.

|  | elpd_diff | se_diff |
|---|---|---|
| chunking * level * mfw_t + (1 + chunking + level + mfw_t \| sample_n) | 0.00 | 0.00 |
| chunking * level * mfw_t + (1 \| sample_n) | -26.10 | 11.83 |
| chunking * level * mfw_t | -55.73 | 17.54 |
| chunking + level + mfw_t + (1 + chunking + level + mfw_t \| sample_n) | -57.04 | 12.14 |
| chunking + level + mfw_t + (1 \| sample_n) | -75.00 | 15.52 |
| chunking + level + mfw_t | -97.92 | 19.30 |

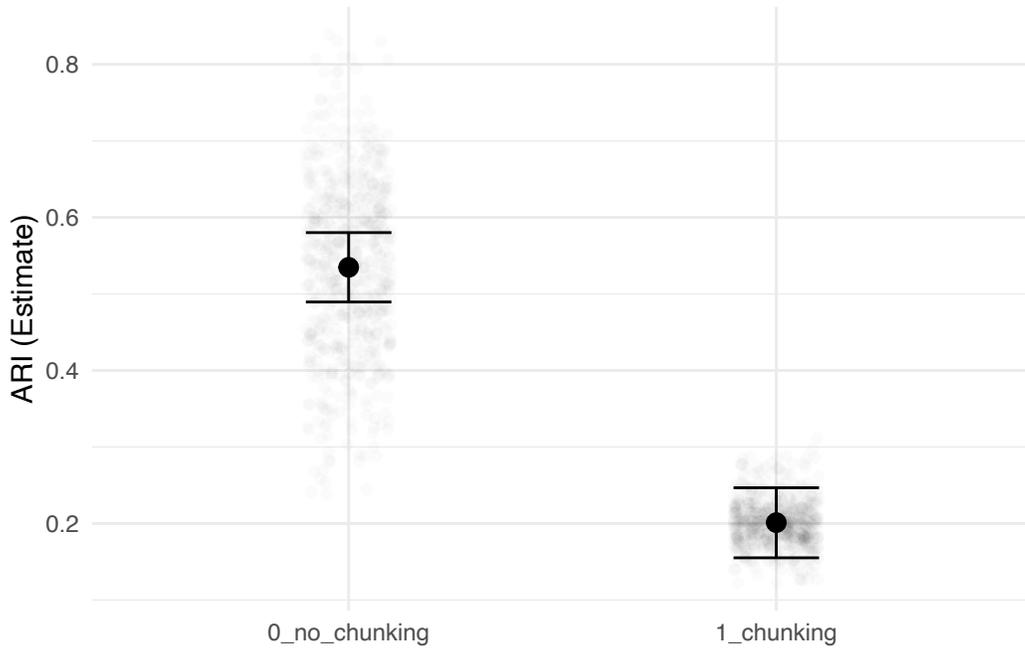

Figure S 22: Effect of chunking on WGCNA performance. Posterior predictions, marginal of samples, superimposed on empirical data points.

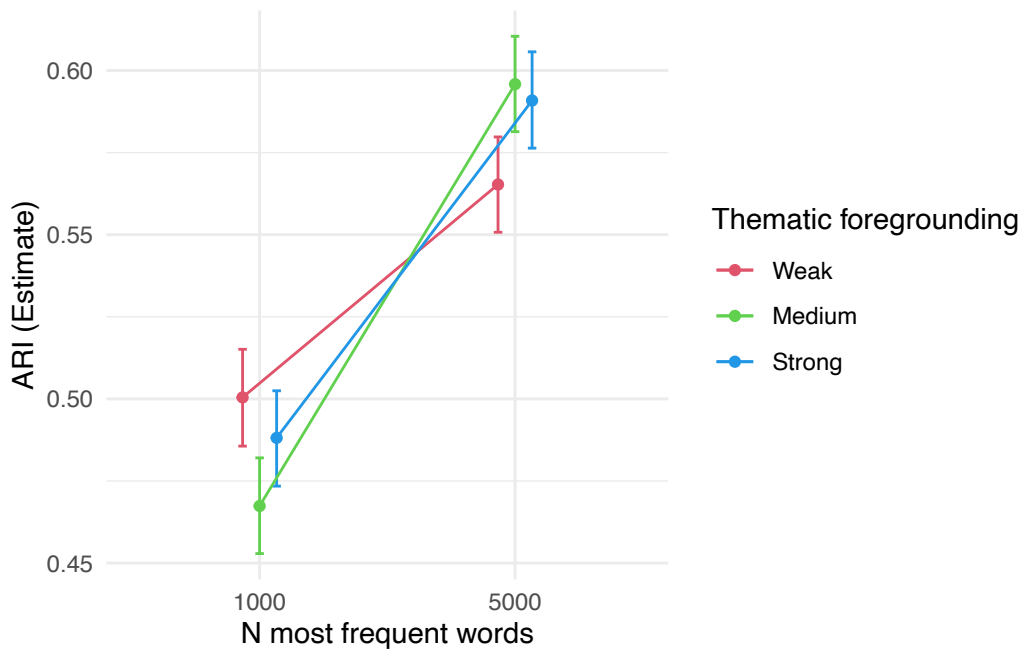

Figure S 23: WGCNA posterior predctions by MFW and thematic foregrounding.

Table S 11: Leave-one-out results for doc2vec group of models.

|  | elpd_diff | se_diff |
|---|---|---|
| ari ~ level + (1 + level \| sample_n) | 0.00 | 0.00 |
| ari ~ level + (1 \| sample_n) | -0.80 | 1.48 |
| ari ~ 1 + level | -18.19 | 6.14 |

### 5.1.8 doc2vec

There is only one predictor for the behavior of doc2vec in our setup: the level of thematic foregrounding. We fit a model with varying slopes per novel sample (Bayesian framework handles single observations in samples just fine):

```
ari ~ 1 + level + (1+level|sample_n)
```

doc2vec embeddings perform similarly across the different levels of thematic foregrounding (Figure S23), which is not surprising, since it uses external representation of semantics and does not depend too much on filtering words. However, there *is* a steady increase in ARI, which means that filtering words and simplifying lexicon *can* improve document representation, even if the same model is used both for semantic similarity scores and document embeddings.

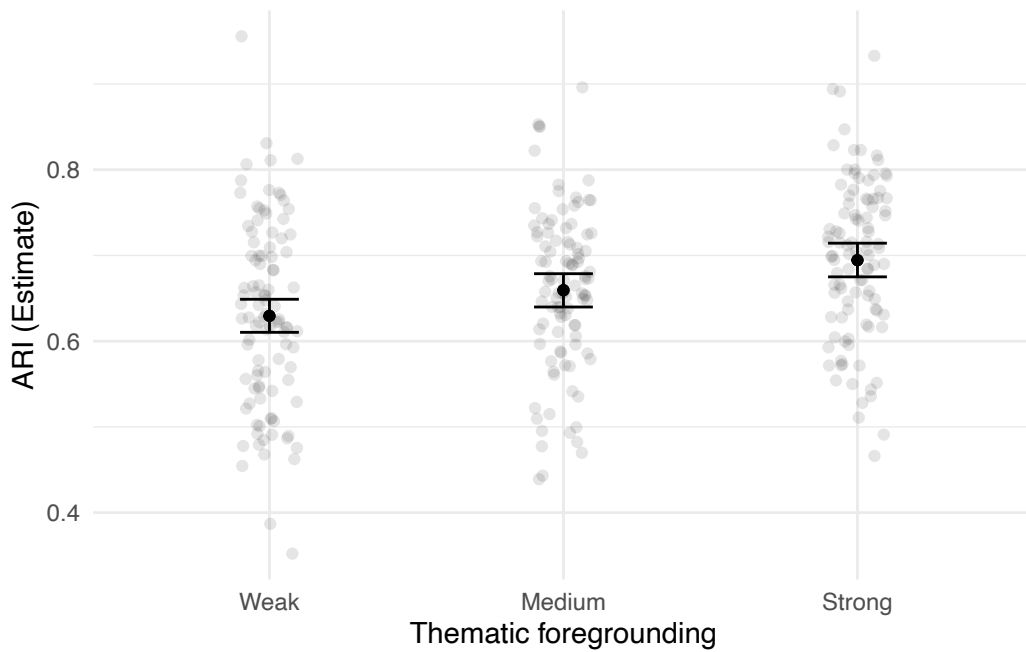

Figure S 24: Posterior predictions for doc2vec performance

## 5.2 Clustering HathiTrust corpus

To test if our results maintain validity in the 'outside' world, we turned to HathiTrust corpus of fiction. We sampled 5000 "unknown" novels from the same period of time (books released after the year 1950). We couldn't just use our small target corpus as a seed of "known" novels, because HathiTrust does not provide original texts: only the token count per page alongside with morphological tagging. It is still possible to train an LDA model with this data, but not reproduce our spaCy pre-processing steps exactly. In addition, many books from our corpus did not have a match in HathiTrust data.

We used another approach. We have found all of the 97 authors from our dataset of four genres in HathiTrust corpus. All the books by these authors were marked as belonging to a corresponding genre. For example, while our original dataset contained only 3 novels by Agatha Christie, HathiTrust contains 71 novels by her. We labeled all of them as "detective" (which, of course, is a simplification). The distribution of books across four genres that we aquire this way is shown on Figure S25. Table S12 shows 10 authors with the largest amount of books.

We chose two combinations of methods to show the difference between 'better' and 'worse' approaches:

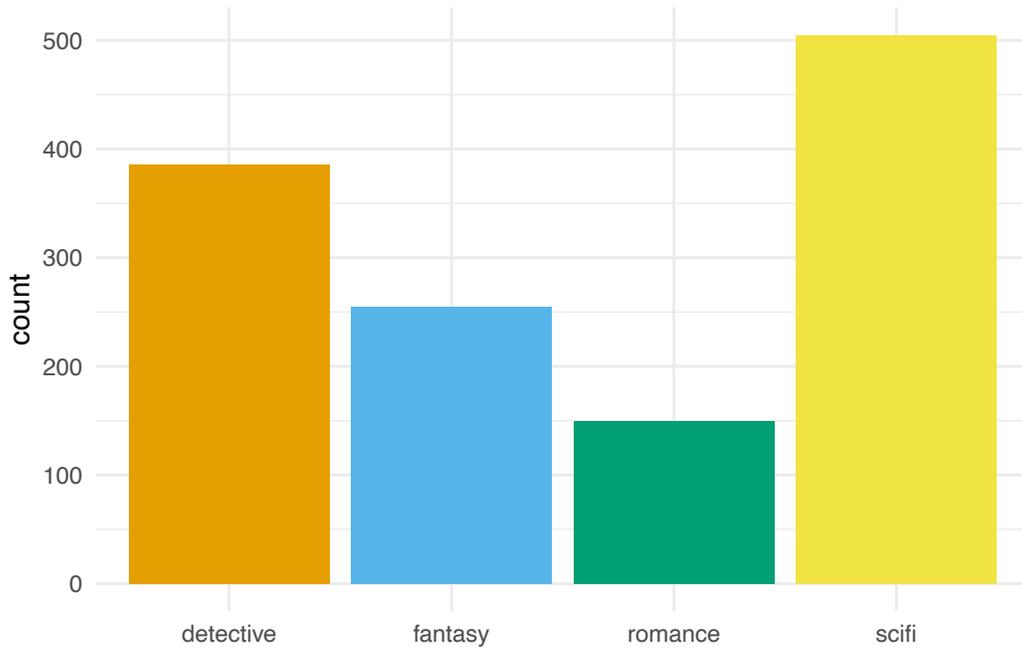

Figure S 25: Genre book counts in HathiTrust data.

Table S 12: Book counts by author.

| genre | author | books |
|---|---|---|
| detective | Christie, Agatha | 71 |
| scifi | Asimov, Isaac | 50 |
| romance | Steel, Danielle | 45 |
| fantasy | Moorcock, Michael | 43 |
| scifi | Silverberg, Robert | 41 |
| scifi | Dick, Philip K | 39 |
| fantasy | Anderson, Poul | 37 |
| scifi | Clarke, Arthur C. (Arthur Charles) | 37 |
| scifi | Aldiss, Brian Wilson | 36 |
| scifi | Heinlein, Robert A. (Robert Anson) | 35 |

1. Better option: LDA model with 1000 MFWs, 100 topics, medium thematic foregrounding, Jensen-Shannon divergence;

2. Worse option: Bag of words, 5000 MFWs, weak thematic foregrounding, cosine distance.

We compare their performance by projecting all 6293 novels in two dimensions with UMAP. We expect a better option to retain visible clusters by genres. Figure S26 sets two projections side by side.

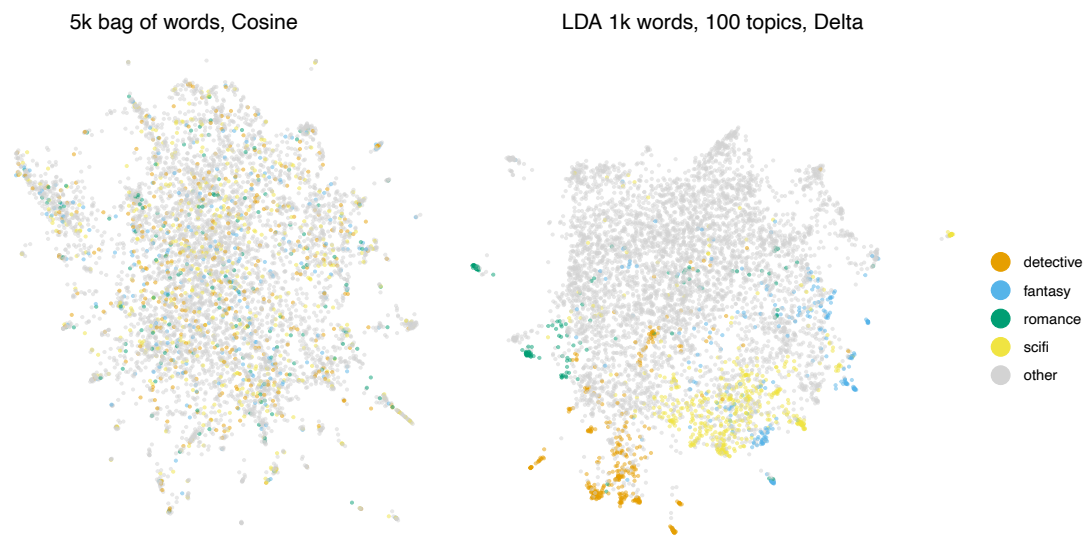

Figure S 26: Two UMAP projections. 'Worse' methods choice on the left, and 'better' methods choice on the right